\documentclass[sigconf, nonacm]{acmart}

\newcommand\vldbdoi{XX.XX/XXX.XX}
\newcommand\vldbpages{XXX-XXX}
\newcommand\vldbvolume{18}
\newcommand\vldbissue{11}
\newcommand\vldbyear{2025}
\newcommand\vldbauthors{\authors}
\newcommand\vldbtitle{\shorttitle} 
\newcommand\vldbavailabilityurl{https://github.com/RUCKBReasoning/OmniSQL}
\newcommand\vldbpagestyle{empty} 

\usepackage{xspace}
\newcommand{\model}{\textsc{OmniSQL}\xspace}
\newcommand{\dataset}{\textsc{SynSQL-2.5M}\xspace}
\newcommand*{\image}[1]{\includegraphics[width=0.25cm,height=!]{#1}}

\usepackage{enumitem}
\usepackage{array}
\usepackage{booktabs}
\usepackage{multirow}
\usepackage{colortbl}
\usepackage{adjustbox}
\usepackage{tcolorbox}
\usepackage{xcolor}

\usepackage{framed}
\definecolor{shadecolor}{rgb}{0.9,0.9,0.9}
\usepackage{balance}

\begin{document}
\title{\model: Synthesizing High-quality Text-to-SQL Data at Scale}

\author{Haoyang Li}
\affiliation{%
  \institution{Renmin University of China}
}
\email{lihaoyang.cs@ruc.edu.cn}

\author{Shang Wu}
\affiliation{%
  \institution{Renmin University of China}
}
\email{wushang\_@ruc.edu.cn}

\author{Xiaokang Zhang}
\affiliation{%
  \institution{Renmin University of China}
}
\email{zhang2718@ruc.edu.cn}

\author{Xinmei Huang}
\affiliation{%
  \institution{Renmin University of China}
}
\email{huangxinmei@ruc.edu.cn}

\author{Jing Zhang}
\affiliation{%
  \institution{Renmin University of China}
}
\email{zhang-jing@ruc.edu.cn}

\author{Fuxin Jiang}
\affiliation{%
  \institution{ByteDance Inc}
}
\email{jiangfuxin@bytedance.com}

\author{Shuai Wang}
\affiliation{%
  \institution{ByteDance Inc}
}
\email{wangshuai.will@bytedance.com}

\author{Tieying~Zhang}
\affiliation{%
  \institution{ByteDance Inc}
}
\email{tieying.zhang@bytedance.com}

\author{Jianjun Chen}
\affiliation{%
  \institution{ByteDance Inc}
}
\email{jianjun.chen@bytedance.com}

\author{Rui Shi}
\affiliation{%
  \institution{ByteDance Inc}
}
\email{shirui@bytedance.com}

\author{Hong Chen}
\affiliation{%
  \institution{Renmin University of China}
}
\email{chong@ruc.edu.cn}

\author{Cuiping Li}
\affiliation{%
  \institution{Renmin University of China}
}
\email{licuiping@ruc.edu.cn}

\thanks{${}^{*}$ Jing Zhang and Tieying Zhang are the corresponding authors.}
\thanks{${}^{\dagger}$ Work done during Haoyang Li's internship at ByteDance.}

\begin{abstract}
Text-to-SQL, the task of translating natural language questions into SQL queries, plays a crucial role in enabling non-experts to interact with databases. While recent advancements in large language models (LLMs) have significantly enhanced text-to-SQL performance, existing approaches face notable limitations in real-world text-to-SQL applications. Prompting-based methods often depend on closed-source LLMs, which are expensive, raise privacy concerns, and lack customization. Fine-tuning-based methods, on the other hand, suffer from poor generalizability due to the limited coverage of publicly available training data. To overcome these challenges, we propose a novel and scalable text-to-SQL data synthesis framework for automatically synthesizing large-scale, high-quality, and diverse datasets without extensive human intervention. Using this framework, we introduce \dataset, the first million-scale text-to-SQL dataset, containing 2.5 million samples spanning over 16,000 synthetic databases. Each sample includes a database, SQL query, natural language question, and chain-of-thought (CoT) solution. Leveraging \dataset, we develop \model, a powerful open-source text-to-SQL model available in three sizes: 7B, 14B, and 32B. Extensive evaluations across nine datasets demonstrate that \model achieves state-of-the-art performance, matching or surpassing leading closed-source and open-source LLMs, including GPT-4o and DeepSeek-V3, despite its smaller size. We release all code, datasets, and models to support further research.
\end{abstract}

\maketitle

\pagestyle{\vldbpagestyle}
\begingroup\small\noindent\raggedright\textbf{PVLDB Reference Format:}\\
\vldbauthors. \vldbtitle. PVLDB, \vldbvolume(\vldbissue): \vldbpages, \vldbyear.\\
\href{https://doi.org/\vldbdoi}{doi:\vldbdoi}
\endgroup
\begingroup
\renewcommand\thefootnote{}\footnote{\noindent
This work is licensed under the Creative Commons BY-NC-ND 4.0 International License. Visit \url{https://creativecommons.org/licenses/by-nc-nd/4.0/} to view a copy of this license. For any use beyond those covered by this license, obtain permission by emailing \href{mailto:info@vldb.org}{info@vldb.org}. Copyright is held by the owner/author(s). Publication rights licensed to the VLDB Endowment. \\
\raggedright Proceedings of the VLDB Endowment, Vol. \vldbvolume, No. \vldbissue\ %
ISSN 2150-8097. \\
\href{https://doi.org/\vldbdoi}{doi:\vldbdoi} \\
}\addtocounter{footnote}{-1}\endgroup

\ifdefempty{\vldbavailabilityurl}{}{
\vspace{.3cm}
\begingroup\small\noindent\raggedright\textbf{PVLDB Artifact Availability:}\\
The source code, data, and/or other artifacts have been made available at \url{https://github.com/RUCKBReasoning/OmniSQL}.
\endgroup
}

\section{Introduction}
Text-to-SQL translates natural language (NL) questions into executable SQL queries, enabling non-experts to interact with databases effectively~\cite{Androutsopoulos1995@nlid, liu2024@nl2sql_survey, li2024@the_dawn}. This capability supports a wide range of data-centric applications and has garnered significant research interest from both natural language processing (NLP) and database communities~\cite{Li2023@resdsql, Pourreza2023@dinsql, Li2023@graphix, Li2024@codes, Gao2024@dailsql, Gu2023@few-shot-text-to-sql, Fu2023@catsql}.

\noindent\textbf{State-of-the-Art: Strengths and Limitations.}
Recent advancements in large language models (LLMs) have driven significant progress in the text-to-SQL field. State-of-the-art (SOTA) solutions~\cite{Gao2024@xiyansql, Pourreza2024@chasesql} often employ multi-agent collaborative frameworks, where specialized agents tackle distinct sub-tasks such as schema linking, text-to-SQL generation, SQL refinement, and SQL selection. Among these, text-to-SQL generation remains the core component. Current approaches to this component mainly rely on LLMs and can be broadly categorized into two paradigms: prompting-based and fine-tuning-based.

Prompting-based methods leverage powerful LLMs through carefully crafted prompts, often relying on closed-source models accessed via APIs. In contrast, fine-tuning-based methods focus on training LLMs on existing text-to-SQL datasets, such as Spider~\cite{Yu2018@spider} and BIRD~\cite{Li2023@bird}, to adapt them for the task. While both approaches have demonstrated impressive benchmark performance, they face notable limitations in real-world applications. Prompting-based methods suffer from challenges such as high usage costs, data privacy concerns, and limited control over model behavior due to their reliance on calling APIs. On the other hand, fine-tuning-based methods often struggle with generalizability to complex problems or domain-specific databases, as publicly available datasets provide limited coverage of real-world scenarios. For instance, experiments reveal that Qwen2.5-Coder-7B-Instruct~\cite{hui2024@qwen2.5-coder}, fine-tuned on the training sets of Spider and BIRD, performs well on their development sets (which share the similar distribution as the training data) but achieves only 43.8\% and 31.4\% execution accuracy on out-of-domain datasets, ScienceBenchmark~\cite{Zhang2023@sciencebenchmark} and EHRSQL~\cite{Lee2022@ehrsql}, respectively. In comparison, zero-shot prompting GPT-4-Turbo~\cite{openai2024@gpt4-turbo} achieves significantly higher execution accuracy of 59.2\% and 43.1\%, highlighting the limited generalizability of current fine-tuning approaches.

To address these limitations, enhancing the text-to-SQL capabilities of open-source LLMs through large-scale, diverse, high-quality training data presents a promising direction. Such efforts could improve both the performance and generalizability of open-source models. Moreover, open-source models can be deployed locally, making text-to-SQL systems more cost-effective, data-secure, and adaptable to modifications, thereby overcoming the challenges associated with closed-source LLMs. However, acquiring large-scale data through human annotation is often infeasible. To mitigate this, several early data augmentation methods~\cite{Yang2021@hierarchical-DA, Hu2023@importance-DA, Wang2021@learning-DA, Wu2021@sql2question-DA} have been proposed to expand existing text-to-SQL datasets. \textit{Unfortunately, most of these methods focus on generating data samples that conform to the distribution of existing datasets, resulting in limited diversity, quality, and scalability. Additionally, many approaches require extensive human effort to define complex templates or grammars, further constraining their practicality.}

\noindent\textbf{Our Proposal.}
To overcome these limitations, we propose a novel text-to-SQL data synthesis framework that distinguishes itself from existing data augmentation methods by offering the following key advantages:
\textbf{(1) Automatic:} The entire synthesis process requires minimal to no human intervention.
\textbf{(2) Scalable:} The framework can generate a large scale of diverse, high-quality data samples, ensuring coverage across a wide range of domains.
\textbf{(3) Realistic:} The synthesized data aligns with real-world user needs and scenarios.

\noindent\textbf{Challenges and Solutions.} Designing an automatic, scalable, and realistic data synthesis framework is a non-trivial task. The primary challenge lies in ensuring automation and scalability while maintaining the quality and diversity of the generated data. To address this, we introduce a progressive pipeline that decouples the synthesis process into several simpler, manageable steps. Each step is automated using LLMs, minimizing human intervention. (1) The pipeline begins by leveraging web tables to synthesize realistic databases. Specifically, given a web table, the LLM is prompted to infer a plausible database business scenario associated with the table and generate a corresponding database. This synthetic database includes multiple relational tables with primary and foreign key relationships. Each relational table contains metadata such as the table name, description, column names, data types, column descriptions, and two example data rows. With millions of web tables available online~\cite{Eggert2023@tablib}, this approach ensures scalability across a wide range of domains. (2) Next, we generate meaningful SQL queries based on the synthesized databases by providing the LLM with the database information. (3) Then, we employ a back-translation technique to convert these SQL queries into semantically equivalent natural language questions. This technique, widely adopted in prior studies~\cite{Hu2023@importance-DA, Awasthi2022@diverse, Guo2018@STAMP, Zhong2020@grounded-DA, Wu2021@sql2question-DA, Zhang2023@sciencebenchmark, Wang2021@learning-DA}, could guarantee the quality of the synthetic <question, SQL query> pairs because converting SQL queries into natural language is inherently more accurate and less ambiguous than the reverse. (4) Finally, to bridge the gap between questions and SQL queries, we draw inspiration from chain-of-thought (CoT) reasoning~\cite{Wei2022@few-shot-cot, Kojima2022@zero-shot-cot}. For each synthetic text-to-SQL data, we generate a step-by-step CoT solution that details the intermediate reasoning steps required to construct the SQL query from the question and the database. This not only enhances the interpretability of the synthetic data but also provides high-quality training signals for text-to-SQL models.

The second challenge is ensuring that the synthetic data aligns with real-world user needs and scenarios. A robust text-to-SQL model must accommodate a wide range of SQL queries, from simple to highly complex, reflecting both basic data retrieval and advanced data analysis requirements. To meet this demand, we define four levels of SQL complexity: simple, moderate, complex, and highly complex. During the SQL synthesis process, we select a complexity level and instruct the LLM to generate SQL queries that correspond to that level. Moreover, real-world users often express their questions in diverse linguistic styles, ranging from formal and explicit to vague and metaphorical. To address this variability, we identify nine common natural language styles: formal, colloquial, imperative, interrogative, descriptive, concise, vague, metaphorical, and conversational. When translating SQL queries into natural language questions, we adopt a specific style and guide the LLM to generate questions consistent with that style. This approach ensures that the synthetic data can accurately capture the various ways users might express their questions in real-world scenarios.

\noindent\textbf{Validation.} To validate the effectiveness of our proposed framework, we introduce \dataset, the first million-scale text-to-SQL dataset. Specifically, \dataset contains 2,544,390 text-to-SQL data samples, each represented as a quadruple of <database, question, SQL query, CoT solution>, spanning 16,583 distinct synthetic databases. Extensive statistics reveal that \dataset demonstrates high diversity and complexity compared to existing text-to-SQL datasets. We further evaluate its quality across four dimensions: database, question, SQL query, and data sample. When compared to the widely adopted human-labeled dataset BIRD~\cite{Li2023@bird}, \dataset outperforms in nearly all criteria, underscoring the reliability and effectiveness of our data synthesis pipeline.

Building on \dataset, we develop \model, a powerful open-source text-to-SQL model available in three scales: 7B, 14B, and 32B. We evaluate \model across nine datasets, including three standard datasets (Spider development and test sets~\cite{Yu2018@spider} and BIRD development set~\cite{Li2023@bird}), three domain-specific datasets (Spider2.0-SQLite~\cite{Lei2024@spider2.0}, ScienceBenchmark~\cite{Zhang2023@sciencebenchmark}, and EHRSQL~\cite{Lee2022@ehrsql}), and three robustness datasets (Spider-DK~\cite{Gan2021@spider-dk}, Spider-Syn~\cite{Gan2021@spider-syn}, and Spider-Realistic~\cite{Deng2021@STRUG}). The results indicate that \model achieves state-of-the-art average performance across these datasets, matching or outperforming leading open-source LLMs (\emph{e.g.}, DeepSeek-V3~\cite{deepseek2024@deepseek-v3} and Qwen2.5-72B-Instruct~\cite{Yang2024@qwen2.5}) and advanced closed-source models (\emph{e.g.}, GPT-4-Turbo~\cite{openai2024@gpt4-turbo} and GPT-4o~\cite{openai2024@gpt4o}), despite its smaller model size. Our contributions are summarized as follows:

\begin{itemize}[leftmargin=1em]
\setlength\itemsep{0em}
    \item \textbf{Data Synthesis Framework.} We propose an automatic and scalable framework for text-to-SQL data synthesis, addressing the generation of realistic databases, complexity-aware SQL queries, stylized natural language questions, and reliable chain-of-thought solutions.
    \item \textbf{Synthetic Dataset and Fine-tuned Multi-scale Model.} We introduce \dataset, the first million-scale text-to-SQL dataset including 2,544,390 diverse and high-quality data samples. Using \dataset, we fine-tune \model, a new text-to-SQL model available in three scales: 7B, 14B, and 32B.
    \item \textbf{New SOTA Text-to-SQL Performance.} Extensive experiments demonstrate that \model achieves new state-of-the-art performance in text-to-SQL tasks, surpassing leading open-source and closed-source LLMs with significantly fewer parameters, highlighting the effectiveness of our data synthesis framework. We have open-sourced our code, datasets, and models on GitHub\footnote{\url{https://github.com/RUCKBReasoning/OmniSQL}} to facilitate further research in text-to-SQL.
\end{itemize}

\section{Related Work}
\subsection{Text-to-SQL}
In the field of text-to-SQL, early studies typically employ an explicit encoder-decoder architecture. The encoder encodes the database schema and the question, while the decoder generates the corresponding SQL query based on the encoded information. Some studies enhance encoders by incorporating graph relation information~\cite{Wang2020@ratsql, Cao2021@lgesql, Li2023@graphix, Cai2021@sadga} or by leveraging pre-training techniques~\cite{Yu2021@grappa, Yin2020@tabert, Deng2021@STRUG}. Other approaches focus on improving decoders by introducing grammar constraints, which help reduce syntax errors in generated SQL queries and thus improve the model's accuracy~\cite{Scholak2021@picard, Wang2018@execution-guided}.

With the rapid advancement of sequence-to-sequence (seq2seq) models, the text-to-SQL task has been transferred to a seq2seq modeling task. Many studies have focused on fine-tuning T5~\cite{Raffel2020@T5} for this purpose~\cite{Li2023@resdsql, Scholak2021@picard, Li2023@graphix, Rai2023@token-preprocessing, Qi2022@rasat}. Recently, the emergence of large language models (LLMs) such as GPT-4~\cite{openai2024@gpt4o, openai2023@gpt4} and Gemini~\cite{Anil2023@gemini, google2024@gemini1.5} has once again transformed the text-to-SQL domain. By leveraging these powerful LLMs, researchers can decompose the complex text-to-SQL task into simpler sub-tasks, including schema linking, text-to-SQL generation, SQL refinement, and SQL selection~\cite{Pourreza2023@dinsql, Gao2024@dailsql, Pourreza2024@chasesql, Gao2024@xiyansql, Wang2025@macsql}. Each sub-task can be managed by an LLM-based agent using prompt engineering or fine-tuning techniques. Unlike previous studies that design complex multi-agent frameworks, this paper focuses on improving the core capability of these frameworks—text-to-SQL generation—using large-scale synthetic data.

\subsection{Data Augmentation for Text-to-SQL}
To address the limited coverage of publicly available datasets, numerous studies have proposed data augmentation methods to generate additional training data (\emph{i.e.}, <question, SQL query> pairs) that align with the distribution of existing datasets. Many approaches~\cite{Kobayashi2025@YORO, Guo2018@STAMP, Zhong2020@grounded-DA, Wu2021@sql2question-DA, Zhang2023@sciencebenchmark, Li2024@codes, Wang2021@learning-DA} employ a ``SQL-to-question'' pipeline, where SQL templates or grammars are used to generate SQL queries, which are then translated into natural language questions using neural network models. This approach ensures high-quality synthetic <question, SQL> pairs, as converting SQL to natural language is generally easier due to the flexibility of natural language. In these methods, SQL templates are often extracted from existing datasets~\cite{Hu2023@importance-DA, Guo2018@STAMP, Zhong2020@grounded-DA, Li2024@codes}, while abstract syntax tree grammars typically require expert design~\cite{Wu2021@sql2question-DA, Zhang2023@sciencebenchmark, Wang2021@learning-DA}. However, SQL templates limit diversity, and grammar-based methods are labor-intensive, making both of them difficult to scale.

Alternatively, some studies adopt a ``question-to-SQL'' pipeline, where questions are first generated and then translated into SQL queries using off-the-shelf text-to-SQL models~\cite{Yang2021@hierarchical-DA}. However, inaccuracies in used text-to-SQL models often result in noisy data. Other methods use question-to-SQL template pairs, either manually crafted or extracted from datasets, to generate new samples by filling slot mappings~\cite{Weir2020@dbpal, Yu2018@SyntaxSQLNet, Yu2021@grappa}. While effective, these approaches suffer from limited diversity and unnatural question generation.

The most closely related work is Sense~\cite{Yang2024@sense}, which employs GPT-4 to directly synthesize new data samples through carefully designed prompts. Specifically, Sense instructs GPT-4 to first generate a database, then formulate a question based on that database, and finally produce the corresponding SQL query for the question. However, Sense's reliance on expensive GPT-4 limits its scalability and cost-effectiveness. In contrast, our framework decouples the synthesis process into simpler, more controllable steps, enabling the use of less powerful yet open-source LLMs. As a result, our approach is highly cost-effective, particularly when scaling to million-level or larger synthesis requirements. Additionally, Sense's ``question-to-SQL'' design risks generating incorrect SQL queries for the previously synthesized questions, whereas our adopted ``SQL-to-question'' strategy ensures higher-quality synthetic data. Beyond these advantages, our method also synthesizes CoT solutions, further enhancing the interpretability of the generated data.

\section{Data Synthesis Framework}
\begin{figure*}[t]
    \centering
    \includegraphics[width=0.85\textwidth]{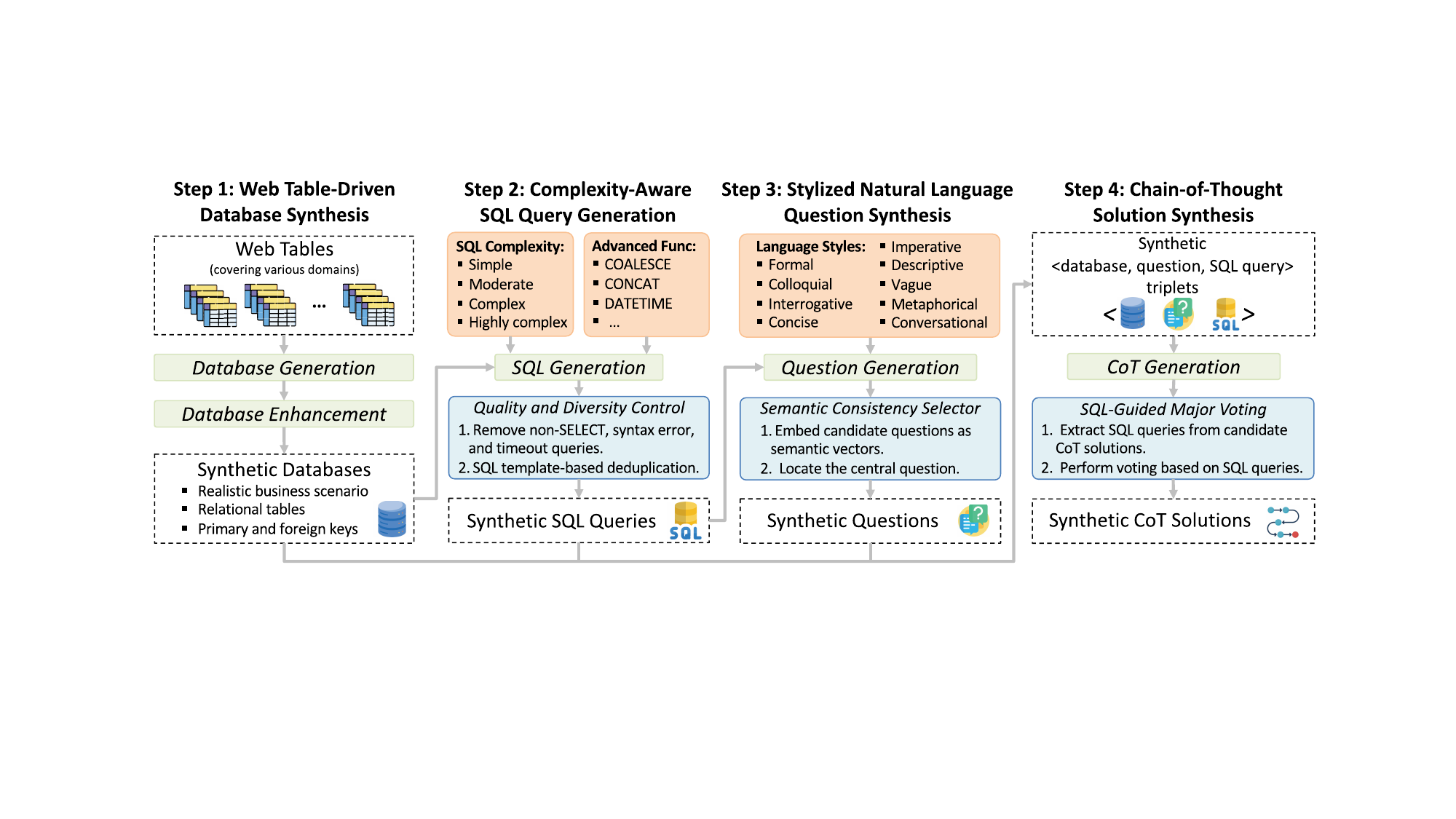}
	\caption{\label{fig:overview} Illustration of the proposed text-to-SQL data synthesis framework.}
\end{figure*}

\subsection{Overview}
As illustrated in Figure~\ref{fig:overview}, our data synthesis framework adopts a progressive pipeline comprising four key steps: web table-driven database synthesis, complexity-aware SQL query generation, stylized natural language question synthesis, and chain-of-thought solution synthesis. Each step leverages large language models (LLMs) in conjunction with automated pre-processing and post-processing strategies to ensure high-quality and diverse outputs, significantly reducing the reliance on extensive human intervention. 

The process begins with web tables, which are abundant on websites and store structural data spanning a wide range of real-world domains. Using these tables, we synthesize databases that emulate realistic business scenarios. Next, based on synthetic databases, we generate SQL queries of varying complexity levels and back-translate them into natural language (NL) questions with diverse language styles. Finally, for each synthetic <database, question, SQL query> triplet, we produce a chain-of-thought (CoT) solution that outlines the step-by-step reasoning process used to derive the SQL query from the natural language question.

\subsection{Web Table-Driven Database Synthesis}\label{sec:database_synthesis}
Developing robust text-to-SQL models requires fine-tuning on diverse databases. However, real-world databases are scarce on the internet because enterprise databases often contain sensitive information. Despite this, we observe that tabular data is abundant~\cite{Eggert2023@tablib, Bhagavatula2015@wikitables, Hulsebos2023@gittables} and also reflects real-world scenarios for structured data storage. This wealth of tabular data presents a unique opportunity to address the aforementioned challenges. Leveraging this, we propose a new database synthesis method comprising two key steps: table-driven database generation and database enhancement. 

\textbf{Table-driven database generation.}  
Specifically, given a web table, we prompt the LLM to first create a realistic business scenario hidden behind the given table and then design a relational database that could store the data relevant to this scenario. Each generated database includes several relational tables, along with structural information such as primary and foreign keys. Each relational table is defined by a table name, a description, column names, column data types, column descriptions, and two example data rows.  

To encourage the synthesis of complex databases, we instruct the LLM to generate $K$ relational tables. In this work, $K$ is an integer sampled from a normal distribution $\mathcal{N}(10, 4^2)$. The prompt used for database generation comprises the following components: \textbf{(1) Task Instruction}: Directs the LLM to generate a business scenario and a corresponding database containing $K$ relational tables, based on the given web table. \textbf{(2) 2-Shot Demonstrations}\footnote{Examples of the 2-shot demonstrations are available at: \url{https://github.com/RUCKBReasoning/OmniSQL/blob/main/data_synthesis/database_synthesis/prompt_templates/schema_prompt.txt}}: Includes two human-crafted examples of database synthesis, each presented as a <web table, business scenario, database> triplet. \textbf{(3) Web Table}: The input web table that serves as the seed for database generation.

\textbf{Database Enhancement.} Although the LLM can generate meaningful and realistic databases by following our instructions, we observe two practical issues with the synthetic databases: overly simplistic relational tables (averaging 4 columns per table) and incomplete primary and foreign key relationships. We attribute these limitations to shortcut learning in LLMs, where models may prioritize fulfilling user instructions with minimal effort rather than achieving optimal complexity or completeness~\cite{Du2022@shortcut-learning-survey}. To address these shortcomings and ensure the synthetic databases meet the complexity requirements of real-world applications, we introduce a database enhancement step following the initial generation.

In this step, given an initially generated database, we prompt the LLM to expand it by adding relevant columns to each relational table and completing any missing primary and foreign key relationships. This enhancement increases the average number of columns per table and improves the structural integrity of the databases. The prompt used for the database enhancement consists of the following components: \textbf{(1) Task Instruction}: Directs the LLM to enhance the structure of the given database.
\textbf{(2) Database Information}: The initially generated database along with its business scenario.

Our database synthesis method is highly scalable, as each web table can be transformed into a database, and millions of web tables are readily available online~\cite{Eggert2023@tablib}. Furthermore, previous work~\cite{Yang2024@sense} has shown that LLMs are already proficient at generating high-quality database schemas, which can be attributed in part to the extensive presence of \texttt{CREATE TABLE} statements in large web-scale corpora. To empirically support this claim, we analyze the pre-training data used for StarCoder~\cite{Li2024@starcoder}, a prominent code-focused LLM, and find 975,420 \texttt{.sql} files. Notably, 42\% of these SQL files contain at least one \texttt{CREATE TABLE} statement. This widespread exposure to database schema definitions during pre-training likely contributes to the strong database generation capabilities observed in LLMs. In contrast, existing methods for collecting large-scale relational databases often struggle with scalability or quality issues. For example, GitSchemas~\cite{Dohmen2022@GitSchemas} and SchemaPile~\cite{Dohmen2024@SchemaPile} extract database construction statements (\emph{e.g.}, \texttt{CREATE TABLE} and \texttt{ALTER TABLE}) from SQL files on GitHub. However, the availability of high-quality SQL files is limited compared to the abundance of web tables, restricting their scalability. WikiDBs~\cite{Vogel2024@wikidbs} constructs databases using web tables from Wikidata. Specifically, it incrementally builds databases by starting with a single table and extending it with related tables. However, this approach risks connecting unrelated tables, potentially compromising database consistency.

\subsection{Complexity-Aware SQL Query Generation}\label{sec:sql_generation}
After generating databases, the next step is to synthesize SQL queries. In this paper, we leverage LLMs for SQL query generation, as their pre-training exposes them to a wide range of SQL scripts and snippets, making them well-suited for the task. However, early experiments revealed a challenge: smaller LLMs tend to generate overly simple SQL queries, while larger LLMs often produce highly complex ones, leading to an imbalance in query complexity. Recent literature~\cite{ma2025@comparing} also reports a similar imbalance in query complexity when employing LLMs to generate SQL queries from the query descriptions of the TPC-DS~\cite{tpcds} benchmark. To address this issue and control the complexity distribution of the synthetic data, we define four complexity levels—simple, moderate, complex, and highly complex—and instruct the LLM to generate SQL queries that align with a specified level. Additionally, we provide the LLM with advanced SQL functions and sampled database values to ensure the generated queries are meaningful and realistic. Furthermore, since many real-world natural language questions seek specific items (\emph{e.g.}, a statistical number or a person’s name), the corresponding SQL queries often return a limited number of columns. To reflect this, we impose a constraint on the number of returned columns during SQL query synthesis. Specifically, the prompt for SQL query synthesis includes the following components:
\begin{itemize}[leftmargin=1em]
\setlength\itemsep{0em}
    \item \textbf{Task Instruction}: Directs the LLM to generate meaningful SQL queries that meet real-world data analysis needs.
    \item \textbf{Database Schema}: Includes the \texttt{CREATE TABLE} statements for all relational tables in the given database.
    \item \textbf{Advanced SQL Functions}: Randomly samples a few advanced SQL functions supported by the database engine, allowing the LLM to incorporate these functions in its queries when appropriate\footnote{This paper focuses on the SQLite engine, as many text-to-SQL benchmarks are based on it. The functions supported by SQLite, including names and descriptions, can be found in the official documentation: \url{https://www.sqlite.org/lang_corefunc.html}.}. Each SQL function is presented with its name and a detailed description to help the LLM properly use it.
    \item \textbf{Database Values}: Randomly samples a few columns along with their stored values to assist the LLM in generating meaningful and contextually relevant predicates.
    \item \textbf{SQL Complexity}: Randomly samples a complexity level from [``Simple'', ``Moderate'', ``Complex'', ``Highly complex'']. Each level is defined by specific criteria along with an example SQL query. 
    \item \textbf{Column Selection Constraint}: Specifies the number of columns the synthetic SQL query must select. This value is sampled from a geometric distribution with a success probability of $p = 0.6$. This distribution is chosen because it naturally biases the sampling toward smaller numbers, reflecting the common text-to-SQL scenario where SQL queries typically select a few columns.
\end{itemize}

In the post-processing stage, we apply several quality control measures. First, we filter out non-SELECT queries using pre-defined rules and execute the remaining queries on the synthetic databases to eliminate those with syntax errors or that result in timeouts. Then, to ensure diversity, we extract SQL templates\footnote{Templates are extracted by masking only the values in SQL queries. For example, given the SQL query ``\texttt{SELECT name FROM school WHERE age > 18}'', its template is ``\texttt{SELECT name FROM school WHERE age > [MASK]}''.} and retain only one query per template. For example, if the LLM generates ``\texttt{SELECT name FROM school WHERE age > 18}'' and ``\texttt{SELECT name FROM school WHERE age > 55}'', which share the same template, only one of these queries is retained in the final dataset.

\subsection{Stylized NL Question Synthesis}
After synthesizing SQL queries, the next step is to translate them into semantically equivalent natural language (NL) questions. Existing studies have primarily focused on ensuring the semantic accuracy of synthetic questions by introducing various novel techniques, such as hierarchical generation~\cite{Wu2021@sql2question-DA}, intermediate representations~\cite{Hu2023@importance-DA}, and pointer-decoder networks~\cite{Zhong2020@grounded-DA}. In this work, we argue that linguistic diversity is equally critical for developing robust text-to-SQL models, as real-world users express their questions in a wide range of styles. This is further supported by recent robustness benchmarks~\cite{Chang2023@dr.spider, Gan2021@spider-syn, Deng2021@STRUG}, which reveal that many text-to-SQL models struggle with linguistic perturbations, such as synonym substitution or sentence paraphrasing. Therefore, we advocate for a dual focus on both semantic accuracy and linguistic diversity during question synthesis.

To enhance linguistic diversity, we define nine language styles commonly observed in real-world user questions: formal, colloquial, imperative, interrogative, descriptive, concise, vague, metaphorical, and conversational. The first six styles (formal, colloquial, imperative, interrogative, descriptive, and concise) reflect scenarios where users express their intentions clearly but with variations in tone. In contrast, the vague and metaphorical styles represent cases where users employ ambiguous vocabulary or figurative language, often requiring external knowledge for interpretation. Finally, the conversational style simulates multi-turn dialogues, where users iteratively clarify their intentions. This style is particularly relevant in real-world applications, as users may not always express their needs directly, necessitating follow-up questions from the model. Examples illustrating these language styles can be found in our open-source GitHub repository\footnote{\url{https://github.com/RUCKBReasoning/OmniSQL/blob/main/assets/example_questions.png}}.

We note that DBPal~\cite{Weir2020@dbpal} also considers linguistic diversity during its question synthesis process. However, it relies on an off-the-shelf
paraphrasing database~\cite{Pavlick2016@ppdb} to replace synonyms in existing questions, which limits its ability to cover the full range of styles commonly used by users. Moreover, this simple synonym replacement mechanism may result in unnatural questions, further highlighting the need for a more robust approach to achieve linguistic diversity.

Specifically, the prompt for synthesizing natural language questions consists of the following components:
\begin{itemize}[leftmargin=1em]
    \setlength\itemsep{0em}
    \item \textbf{Task Instruction}: Directs the LLM to first generate an explanation of the provided SQL query and then translate it into a natural language question.  
    \item \textbf{SQL Query}: The SQL query to be translated. 
    \item \textbf{SQL-related Column Information}: Includes the names and descriptions of columns referenced in the SQL query. This aids the LLM in generating semantically accurate questions, particularly when column names are ambiguous, abbreviated, or coded.
    \item \textbf{Desired Language Style}: A randomly sampled style from the nine predefined styles, each accompanied by a description and an example question. For the formal, colloquial, imperative, interrogative, descriptive, and concise styles, the LLM generates stylized questions directly. For the vague and metaphorical styles, the LLM additionally provides the external knowledge underlying the question. For the conversational style, the LLM generates a multi-turn dialogue between <User> and <Assistant>.
\end{itemize}

For each synthetic SQL query, we generate multiple candidate questions using the LLM. To select the most semantically accurate question, we introduce a semantic consistency selector module, inspired by~\cite{Zhang2023@sciencebenchmark, Rossiello2017@centroid}. Specifically, we utilize Sentence Transformers~\cite{Reimers2019@sentencebert} to embed the candidate questions into vector representations\footnote{We use the ``all-mpnet-base-v2'' model for sentence embedding due to its superior embedding quality. Further details are available at \url{https://sbert.net/docs/sentence_transformer/pretrained_models.html}.}. For each candidate question, we compute its average cosine similarity with all other candidates. The question with the highest average similarity is selected, as it lies closest to the semantic center of the candidate set. By combining the pre-defined language styles with the semantic consistency selector, we enhance both the linguistic diversity and semantic accuracy of the synthetic questions.

\begin{table*}[t]
    \centering
    \caption{Overall statistics of different datasets. Note: $\dagger$ denotes that each database in WikiSQL consists of a single table. $*$ indicates that the number of unique SQL queries for BIRD cannot be calculated due to the inaccessibility of its test set.}
    \label{tab:overall_statistics}
    \setlength{\tabcolsep}{8pt}
    \footnotesize
    \renewcommand{\arraystretch}{0.9}
    \begin{tabular}{l|ccccccc}
       \toprule
       \textbf{Dataset} &  \textbf{Source} & \textbf{\# Example} & \textbf{\# Unique SQL} & \textbf{\# DB} & \textbf{Lang. Styles} & \textbf{Knowledge} & \textbf{CoT Solution} \\
       \midrule
       WIKISQL~\cite{Zhong2017@wikisql} & Human+Template & 80,654 & 80,257 & 26,531$^{\dagger}$ & \image{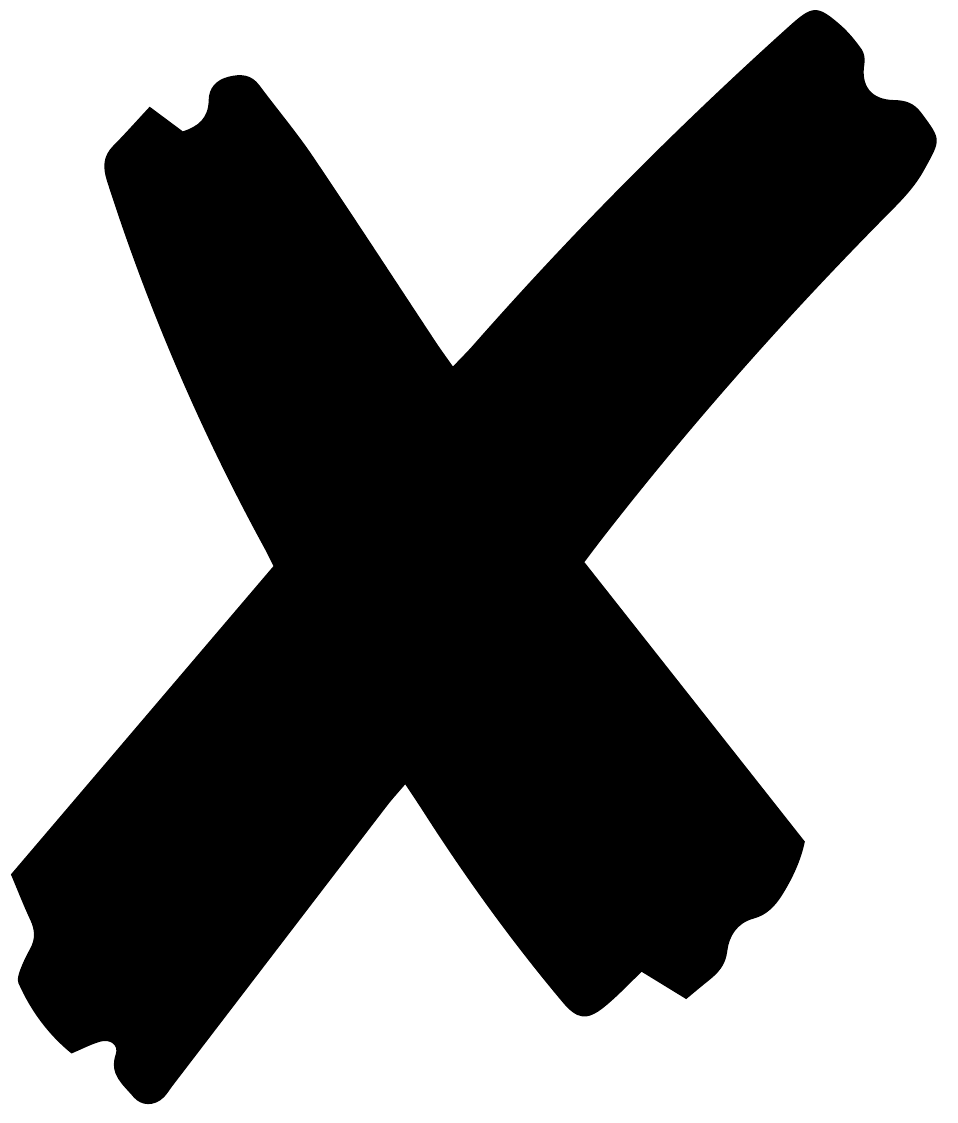} & \image{figures/not} & \image{figures/not} \\
       Spider~\cite{Yu2018@spider} & Human & 10,181 & 4,489 & 200 & \image{figures/not} & \image{figures/not} & \image{figures/not} \\
       BIRD~\cite{Li2023@bird} & Human & 12,751 & -$^{*}$ & 95 & \image{figures/not} & \image{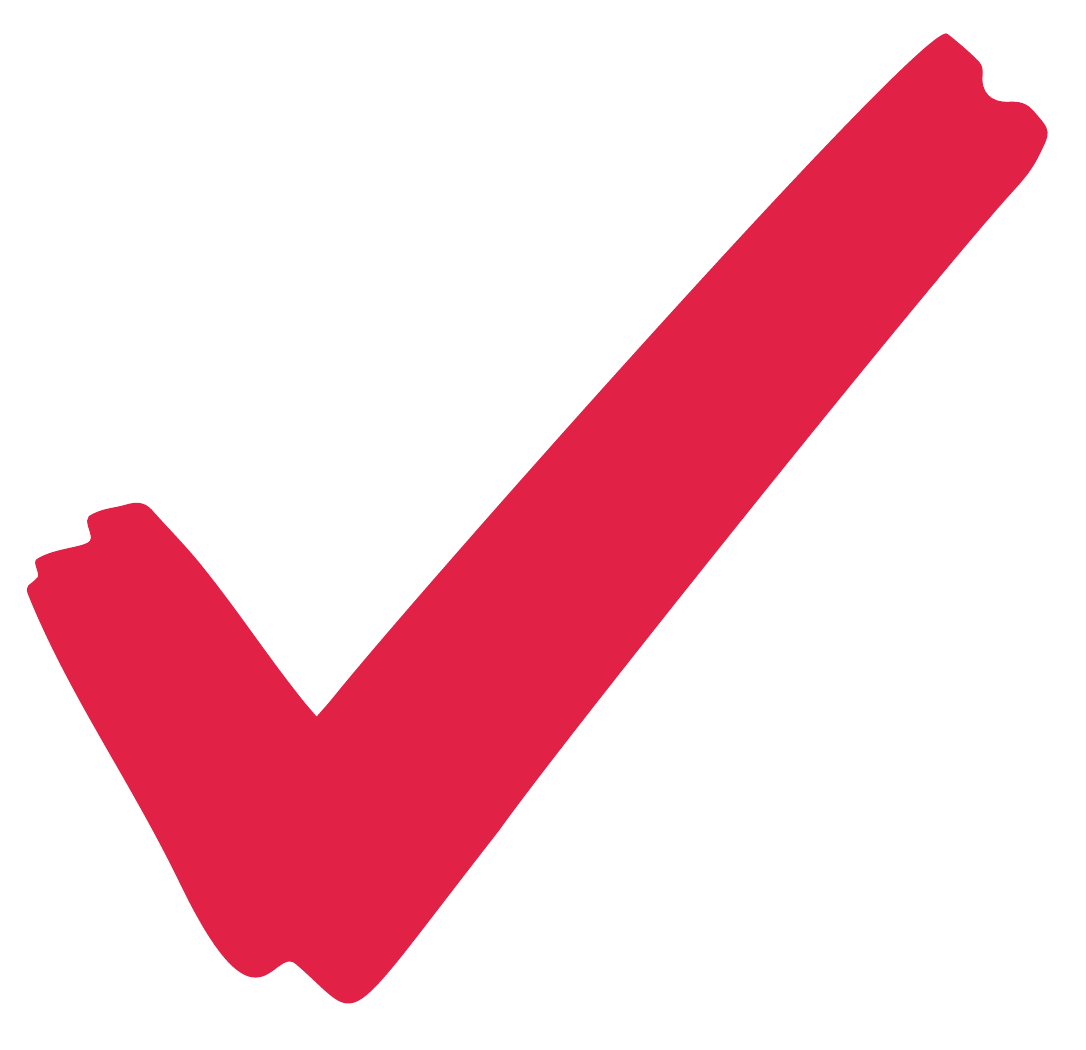} & \image{figures/not} \\
       ScienceBenchmark~\cite{Zhang2023@sciencebenchmark} & LLM-Gen+Human+Template & 5,031 & 3,652 & 3 & \image{figures/not} & \image{figures/not} & \image{figures/not} \\
       EHRSQL~\cite{Lee2022@ehrsql} & Human+Template & 20,108 & 18,253 & 2 & \image{figures/not} & \image{figures/not} & \image{figures/not} \\
       \midrule
       \dataset & LLM-Gen & 2,544,390 & 2,412,915 & 16,583 & \image{figures/yes} & \image{figures/yes} & \image{figures/yes} \\
    \bottomrule
    \end{tabular}
\end{table*}

\subsection{Chain-of-Thought Solution Synthesis}\label{sec:cot_synthesis}
Chain-of-thought (CoT) reasoning has demonstrated remarkable success across various challenging tasks~\cite{Wei2022@few-shot-cot, Kojima2022@zero-shot-cot}. By decomposing complex problems into smaller, manageable steps, this approach enables LLMs to tackle intricate tasks more effectively, improving both accuracy and interpretability. Building on this, we augment the synthetic <database, question, SQL query> triplets by generating CoT solutions that explicitly outline the reasoning process behind constructing the SQL query from the question. The prompt for synthesizing CoT solutions consists of the following components:
\begin{itemize}[leftmargin=1em]
    \setlength\itemsep{0em}
    \item \textbf{Task Instruction}: Directs the LLM to generate a step-by-step CoT solution using the provided information.
    \item \textbf{Database Schema}: Includes the \texttt{CREATE TABLE} statements for all relational tables in the database.
    \item \textbf{NL Question and SQL Query Pair}: The natural language question and its corresponding SQL query.
\end{itemize}

A typical CoT solution begins by analyzing the question to identify the key information required. It then determines the relevant tables, columns, and filtering criteria needed to retrieve the desired data. Finally, it constructs the SQL query step by step, incorporating necessary joins, filters, aggregations, groupings, and other operators, culminating in the complete SQL query as the final answer. 

Interestingly, in our preliminary experiments, we observe that the SQL queries generated by the synthetic CoT sometimes differ from the original ones. Upon closer examination, we find that CoT-generated SQL queries often better align with the questions compared to the original SQL queries. This improvement arises because the original <database, question, SQL query> triplets occasionally contain minor issues, such as unnecessary column selections, and incorrect join paths. The CoT synthesis process allows the LLM to identify and correct these issues during step-by-step reasoning, resulting in more accurate and refined SQL queries. This observation also aligns with prior research showing that LLMs excel at detecting and resolving minor errors in predicted SQL queries~\cite{Gao2024@xiyansql, Pourreza2024@chasesql, Talaei2024@chess}. Thus, incorporating CoT not only provides detailed solutions but also enhances the overall quality of the synthetic data.

To enhance the diversity and reliability of synthetic CoT solutions, we generate multiple candidate CoT solutions for each synthetic <database, question, SQL query> triplet. To select the most reliable CoT solution, we extract SQL queries from these candidates and perform a majority vote. Specifically, we group candidates based on the execution results of their SQL queries. The final CoT solution is selected from the group with the most votes.
\section{\dataset: A Million-Scale Dataset}

To demonstrate the effectiveness of our data synthesis framework, we introduce \dataset—the first million-scale text-to-SQL dataset, entirely generated automatically by LLMs. \dataset contains 2,544,390 high-quality text-to-SQL samples, each represented as a <database, question, SQL query, CoT solution> quadruple. The dataset spans 16,583 synthetic databases across a wide range of real-world domains, including social media sentiment analysis, product inventory management, movie analytics, galaxy morphological analysis, and more.

To mitigate potential bias introduced by the habits of specific LLMs, we employ multiple LLMs during the data synthesis process, with each model responsible for generating a portion of the data. A key advantage of our framework is its decomposition of the text-to-SQL data synthesis task into four simple and manageable sub-tasks. Each sub-task can be effectively handled by relatively smaller, open-source LLMs, which are locally deployable and cost-efficient. This design eliminates the need for heavy reliance on expensive closed-source LLMs, as seen in prior studies~\cite{Yang2024@sense}. In practice, we utilize models from multiple open-source model families, including Llama3.1~\cite{Dubey2024@llama3} (Meta-Llama-3.1-8B/70B-Instruct), Deepseek Coder~\cite{Guo2024@deepseek-coder, deepseek@deepseek-coder-v2} (DeepSeek-Coder-6.7B/33B-Instruct, DeepSeek-Coder-V2-Lite-Instruct), Qwen2.5~\cite{Yang2024@qwen2.5} (Qwen2.5-7B/14B/32B/ 72B-Instruct), and Qwen2.5 Coder~\cite{hui2024@qwen2.5-coder} (Qwen2.5-Coder-7B/14B/32B-Instruct). Larger models are assigned a greater share of the synthesis workload to ensure data quality.

To construct \dataset, we sample 0.1\% of data from a tabular corpus, TabLib~\cite{Eggert2023@tablib}, resulting in approximately 1,319,561 web tables. Since these tables are sourced from websites, they often contain incomplete, redundant, or irrelevant content, which could compromise the quality of synthetic databases. To address this, we design a systematic filtering pipeline with the following steps: (1) Language Filtering: Non-English tables are removed to align with the English benchmarks. (2) Size Filtering: Tables with fewer than 5 columns or 5 rows are eliminated. (3) Deduplication: Similar tables are removed based on table headers. (4) Semantic Evaluation: Qwen2.5-72B-Instruct~\cite{Yang2024@qwen2.5} is used to assess and filter out tables with insufficient semantic richness. After filtering, 19,935 high-quality web tables are retained. During the database synthesis process, 16,583 tables are successfully expanded into structurally complete databases, despite encountering parsing errors in LLM-generated responses and database creation failures. The initially generated databases contain an average of 9.15 tables per database and 4.86 columns per table. After enhancement, these averages increase to 10.15 tables and 7.3 columns, respectively, resulting in databases with real-world complexity. All databases are hosted and managed using SQLite. In the SQL synthesis phase, we generate 300 SQL queries for each synthetic database, producing approximately 5 million synthetic SQL queries. After applying quality and diversity controls, around 2.5 million queries are retained. For the natural language question and CoT synthesis phase, we sample 8 responses from LLMs for each input prompt, using a temperature of 0.8.

In this section, we provide a comprehensive statistical analysis of \dataset, highlighting its quality, diversity, and complexity through comparisons with three widely-used standard datasets (WikiSQL~\cite{Zhong2017@wikisql}, Spider~\cite{Yu2018@spider}, and BIRD~\cite{Li2023@bird}) and two domain-specific datasets (ScienceBenchmark~\cite{Zhang2023@sciencebenchmark} and EHRSQL~\cite{Lee2022@ehrsql}). Additionally, we evaluate the quality of \dataset using the ``LLM-as-a-judger'' approach, further underscoring its high overall quality.

\subsection{Overall Statistics}
Table~\ref{tab:overall_statistics} provides a comparative overview of \dataset and other text-to-SQL datasets. As the statistics demonstrate, \dataset is a large-scale, cross-domain text-to-SQL dataset that encompasses a diverse range of unique SQL queries and databases. Unlike previous datasets, which rely heavily on human annotators, \dataset is entirely generated by LLMs, showcasing its high scalability and cost efficiency. Furthermore, \dataset incorporates diverse linguistic styles in its questions. For vague or metaphorical questions, external knowledge is synthesized to clarify their intent. Notably, to the best of our knowledge, \dataset is the first large-scale dataset to provide step-by-step CoT solutions. 

\begin{table}[t]
    \centering
    \caption{Database statistics. The asterisk(*) means BIRD’s database statistics are based on training and development sets due to the test set's inaccessibility. ``DE'' is the abbreviation of ``database enhancement''.}
    \label{tab:db_statistics}
    \setlength{\tabcolsep}{3.8pt}
    \footnotesize
    \renewcommand{\arraystretch}{0.9}
    \begin{tabular}{l|ccccc}
       \toprule
       \textbf{Dataset} & \textbf{\# DB} & \textbf{\# Tbl/DB} & \textbf{\# Col/DB} & \textbf{\# PK/DB} & \textbf{\# FK/DB} \\
       \midrule
       WIKISQL~\cite{Zhong2017@wikisql} & 26,531 & 1.00 & 6.34 & 1.00 & 0.00 \\
       Spider~\cite{Yu2018@spider} & 200 & 5.11 & 26.82 & 4.70 & 4.79 \\
       BIRD$^{*}$~\cite{Li2023@bird} & 95 & 7.64 & 54.56 & 6.71 & 6.58 \\
       ScienceBenchmark~\cite{Zhang2023@sciencebenchmark} & 3 & 16.67 & 86.67 & 14.33 & 20.33 \\
       EHRSQL~\cite{Lee2022@ehrsql} & 2 & 13.50 & 92.00 & 13.50 & 17.00 \\
       \dataset (w/o DE) & 16,583 & 9.15 & 44.47 & 9.15 & 8.56  \\
       \midrule
       \dataset & 16,583 & 10.15 & 74.07 & 10.14 & 9.61  \\
    \bottomrule
    \end{tabular}
\end{table}
\begin{table*}[t]
    \centering
    \caption{SQL statistics. ``Agg.'' and ``Func.'' are abbreviations of aggregations and functions. The asterisk(*) means BIRD's SQL statistics are based on training and development sets due to test set inaccessibility. }
    \label{tab:sql_statistics}
    \footnotesize
    \setlength{\tabcolsep}{5pt}
    \renewcommand{\arraystretch}{1.0}
    \begin{tabular}{l|ccccccccccc}
       \toprule
      \multirow{2}{*}{Dataset} & \# Tables & \# Joins & \# Func. & \# Tokens & \multirow{2}{*}{\# Agg.} & \# Set & \multirow{2}{*}{\# Subqueries} & \# Window & \multirow{2}{*}{\# CTEs}  & \# Unique &\# Unique \\
      & per SQL & per SQL & per SQL & per SQL &  & Operators &  & Func. & & Skeletons & Func. \\
       \midrule
        WIKISQL~\cite{Zhong2017@wikisql} & 1.00 & 0 & 0.72 & 10.32 & 22,707 & 0 & 0 & 0 & 0 & 488 & 5 \\
        Spider~\cite{Yu2018@spider} & 1.52 & 0.48 & 0.51 & 14.85 & 3,802 & 348 & 604 & 0 & 0 & 2,136 & 5 \\
        BIRD$^{*}$~\cite{Li2023@bird} & 1.98 & 0.94 & 0.67 & 24.75 & 5,144 & 25 & 843 & 10 & 9 & 4,596 & 24 \\
        ScienceBenchmark~\cite{Zhang2023@sciencebenchmark} & 1.42 & 0.42 & 0.25 & 14.61 & 1,234 & 2 & 23 & 0 & 0 & 855 & 5 \\
        EHRSQL~\cite{Lee2022@ehrsql} & 3.60 & 0.62 & \textbf{2.63} & 41.62 & 11,673  & 166 & 17,989 & 3,207 & 0 & 2,447 & 11\\
        \midrule
        \dataset & \textbf{4.00} & \textbf{1.75} & 1.49 & \textbf{57.05} & \textbf{1,897,440} & \textbf{363,472} & \textbf{612,647} & \textbf{33,916} & \textbf{1,073,483} & \textbf{2,190,988} & \textbf{83} \\
    \bottomrule
  
    \end{tabular}
\end{table*}

\begin{figure}[t]
    \centering
    \includegraphics[width=0.475\textwidth]{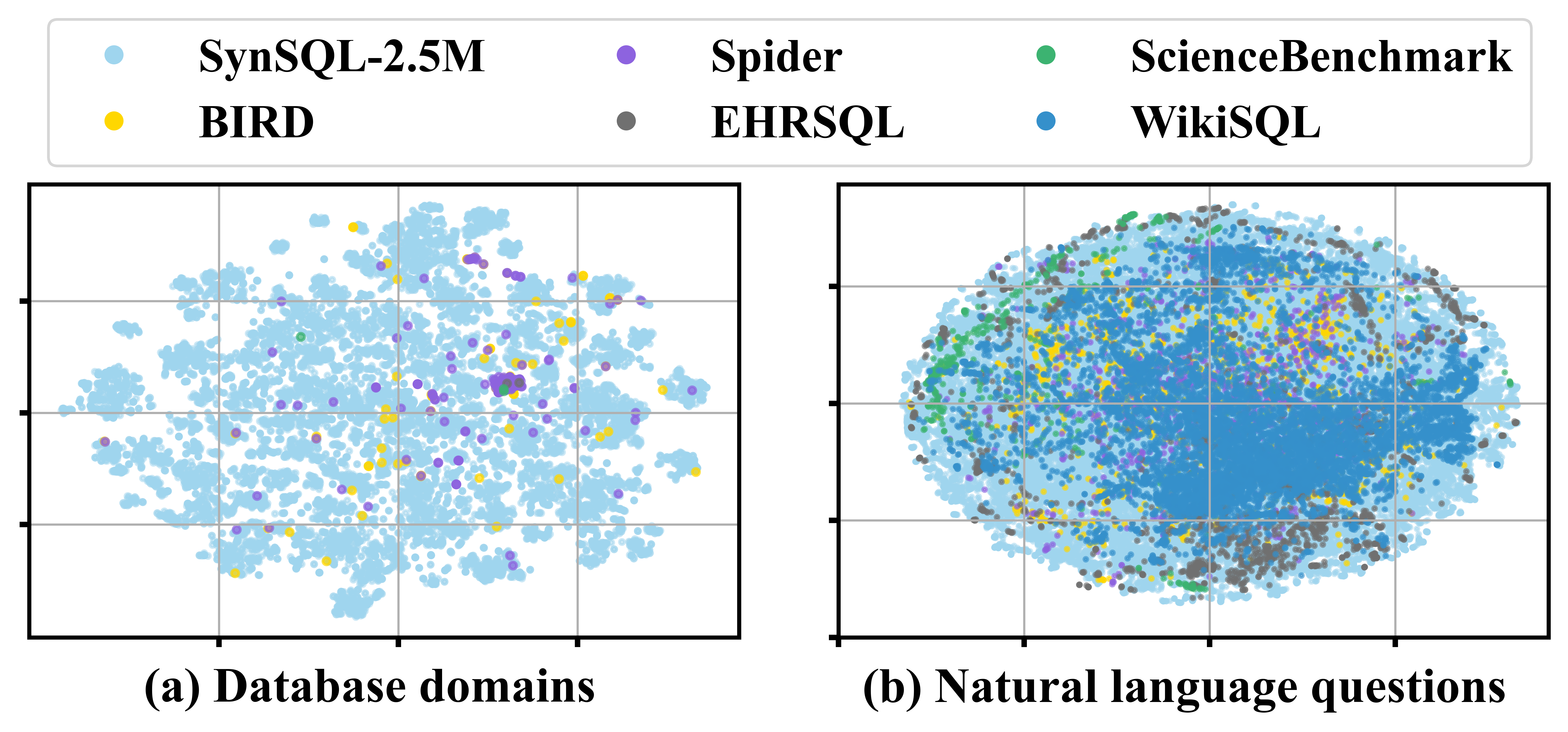}
	\caption{\label{fig:tsne_visualization} Database domain and question visualizations.}
\end{figure}

\subsection{Database Statistics}
\textbf{Database Complexity.} Table~\ref{tab:db_statistics} compares the complexity of our synthetic databases and human-collected databases from existing benchmarks, focusing on the average number of tables, columns, primary keys, and foreign keys per database. The results show that the complexity of our synthetic databases surpasses that of the three widely-used standard datasets (\emph{i.e.}, WikiSQL, Spider, and BIRD). This demonstrates that LLMs can reliably synthesize databases that closely mirror real-world complexity.

\noindent \textbf{Domain Visualization.} Database names (\emph{e.g.}, grain\_data\_analysis and music\_album\_analytics\_and\_management) often reflect the domain of their underlying data. To evaluate the domain diversity of synthetic databases, we leverage Sentence Transformers~\cite{Reimers2019@sentencebert} to embed database names into semantic vectors. In this vector space, databases from similar domains are positioned closer together, while those from dissimilar domains are farther apart. Figure~\ref{fig:tsne_visualization} (a) presents a t-SNE~\cite{Van2008tsne} visualization of these embeddings, highlighting that our synthetic databases exhibit significantly broader domain coverage compared to existing human-collected databases.

\subsection{SQL Statistics}\label{sec:sql_stat}
Table~\ref{tab:sql_statistics} provides a detailed comparison of synthetic SQL queries and existing datasets across 11 dimensions, showcasing the complexity, sophisticated features, and diversity of our synthetic SQL queries.

To assess SQL complexity, we analyze the number of tables, joins, functions, and tokens per SQL query\footnote{SQL queries are tokenized based on whitespace.}. The results reveal that our synthetic SQL queries are more complex than those in human-annotated datasets. For instance, synthetic queries average 1.75 joins per query, compared to 0.94 in BIRD and 0.48 in Spider.

To evaluate the sophisticated features, we calculate the number of SQL queries involving aggregation functions, set operators, subqueries, window functions, and common table expressions (CTEs). The results demonstrate that \dataset provides strong coverage of these advanced features, supporting complex query patterns. Notably, approximately 40\% of synthetic SQL queries utilize CTEs, which enhance the organization and readability of complex queries, making them easier for developers to understand and maintain.

Finally, we assess SQL diversity by counting the number of unique skeletons\footnote{Skeletons are extracted by masking all tables, columns, and values in SQL queries. For example, given the SQL query ``\texttt{SELECT name FROM school WHERE age > 18}'', its skeleton is ``\texttt{SELECT [MASK] FROM [MASK] WHERE [MASK] > [MASK]}''.} and unique functions. The results highlight the high diversity of synthetic SQL queries, with over 2 million unique SQL skeletons. Additionally, synthetic queries cover 83 unique functions supported by the database engine.

\subsection{Question Statistics}
To evaluate the diversity of synthetic questions in \dataset, we employ Sentence Transformers to embed questions into semantic vectors and visualize these embeddings using t-SNE. For clarity, we randomly sample 0.5 million synthetic questions from \dataset for visualization. As illustrated in Figure~\ref{fig:tsne_visualization} (b), \dataset demonstrates extensive coverage, effectively encompassing the distributions of human-annotated datasets.

\subsection{Data Quality Evaluation}
\subsubsection{GPT-4o as Evaluators}
\begin{figure}[t]
    \centering
    \includegraphics[width=0.95\linewidth]{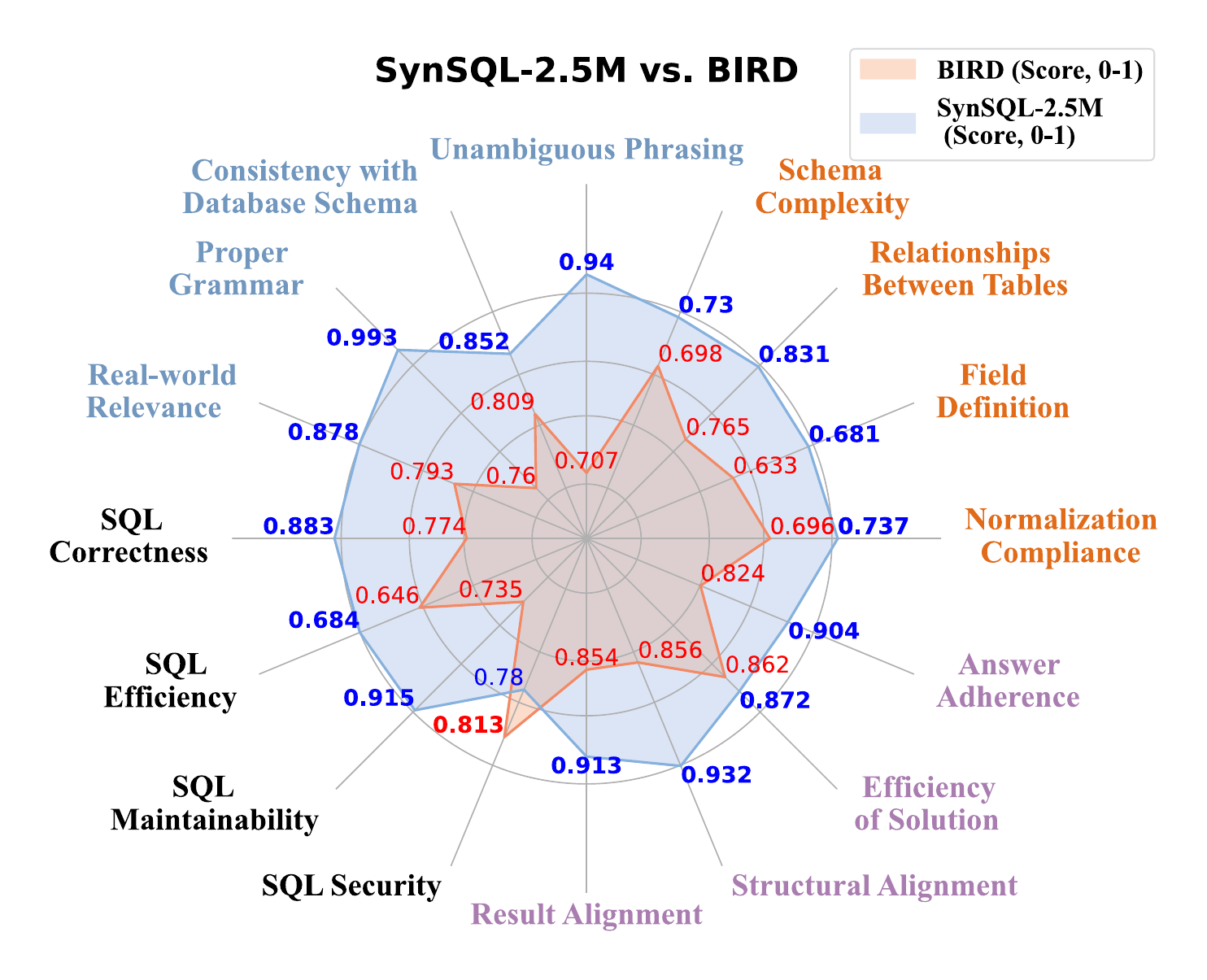}
	\caption{\label{fig:dataset_quality} Quality evaluation of \dataset and BIRD judged by GPT-4o. Scores are computed using Equation~\ref{eq:score}.}
\end{figure}

We comprehensively evaluate \dataset using GPT-4o, a state-of-the-art LLM, following the paradigm of recent LLM-as-a-judge studies~\cite{Zheng2023@llm-as-a-judge, Gu2024@llm-judge-survey, Zhu2023@judgelm}. The evaluation encompasses four key aspects: \textbf{Database Aspect} (normalization compliance, field definition, relationships between tables, and schema complexity), \textbf{Question Aspect} (unambiguous phrasing, consistency with database schema, proper grammar, and real-world relevance), \textbf{SQL Query Aspect} (correctness, efficiency, maintainability, and security), and \textbf{Data Sample Aspect} (result alignment, structural alignment, efficiency of solution, and answer adherence). Detailed descriptions of each criterion can be found in our GitHub repository\footnote{\url{https://github.com/RUCKBReasoning/OmniSQL/tree/main/GPT_evaluation/prompts}}. GPT-4o then assigns a rating (excellent, good, average, or poor) for every criterion on each data sample, along with detailed explanations. We aggregate the results using a weighted average:
\begin{equation}  
    \label{eq:score}
    Score = \frac{N_{e} \times 1.0 + N_{g} \times 0.75 + N_{a} \times 0.5 + N_{p} \times 0.25}{N_{e} + N_{g} + N_{a} + N_{p}},  
\end{equation}  
where \(N_{e}\), \(N_{g}\), \(N_{a}\), and \(N_{p}\) are the numbers of samples rated as excellent, good, average, and poor. For comparison, we also conduct the same evaluation on BIRD~\cite{Li2023@bird}, a widely used human-annotated benchmark. To ensure fairness, we randomly select 1,000 samples from each of \dataset and BIRD for evaluation. As shown in Figure~\ref{fig:dataset_quality}, \dataset outperforms BIRD across nearly all criteria.

\subsubsection{Human Experts as Evaluators}\label{sec:human_eval} To further validate data quality, we conduct a human evaluation on 1,000 sampled data points. The evaluation is performed by three senior graduate students with backgrounds in computer science and active research interests in text-to-SQL and AI4DB. To reduce evaluator workload, we streamline the criteria: human experts only need to provide a binary judgment (pass/fail) for each of the following aspects—database realism and completeness, question meaningfulness, SQL query appropriateness, and overall correctness of the data sample.

To ensure diversity among the evaluated samples, we first randomly select 1,000 synthetic databases. For each database, we then randomly sample one complete text-to-SQL data instance, which includes a database schema, a natural language question, an SQL query, and its corresponding chain-of-thought (CoT) solution. This process results in 1,000 distinct samples for evaluation.

Results show that 96\% of synthetic databases are judged as realistic and structurally complete, making them suitable for practical use. Among the 1,000 complete text-to-SQL samples, 97\% of questions are meaningful, 89\% of SQL queries are appropriate, and 86\% of the complete text-to-SQL data samples are fully correct. These findings indicate that the majority of our synthetic data is high-quality silver-standard data, suitable for model training.
\section{\model: State-of-the-Art Open-Source Text-to-SQL LLM}\label{sec:omnisql}
Using our synthetic dataset, \dataset, along with two widely-adopted datasets, Spider and BIRD, we present \model, a powerful text-to-SQL model available in three scales: 7B, 14B, and 32B. In this section, we describe the construction of input-output sequence pairs for training \model and outline its training objectives. Additional implementation details are provided in Section~\ref{sec:impl_details}.

\subsection{Input-Output Construction}
For all text-to-SQL data samples in the dataset, we first convert them into input-output sequence pairs to facilitate the training of \model. Specifically, the input sequence consists of the database schema and the natural language question. Following previous studies~\cite{Rajkumar2022@eval-llm-text2sql, Yang2024@sense}, the database schema is formatted as \texttt{CREATE TABLE} statements. Additionally, inspired by prior work~\cite{Li2024@codes, Talaei2024@chess}, we enrich the input with three supplementary elements: column descriptions, representative values, and question-relevant values, which are included as comments for each column.

Specifically, column descriptions assist the LLM in identifying the correct columns referenced in the question, particularly when column names are ambiguous (\emph{e.g.}, abbreviations). Representative values for each column inform the LLM about the format of stored values, aiding in the use of advanced functions like \texttt{CONCAT}, \texttt{STRFTIME}, and \texttt{SUBSTR} in the generated SQL query. In practice, we select two distinct values for each column. Finally, following~\cite{Li2024@codes}, we extract question-relevant values from the database, which can help the LLM to generate accurate predicates, especially when the question mentions specific database values. 

The output sequence is the CoT solution, which includes step-by-step reasoning process and the final SQL query. However, since Spider and BIRD only provide gold SQL queries without CoT solutions, we enhance their training sets by synthesizing CoT solutions using the technique described in Section~\ref{sec:cot_synthesis}.

\subsection{Supervised Fine-Tuning}
We fine-tune the LLM using a conditional next token prediction loss, as described in~\cite{Radford2018@gpt1}. The loss function is defined as follows:
\begin{equation}
\text{Loss} = - \mathbb{E}_{(x, y) \sim D} \left[\textstyle \sum_{t} \log P_{\theta}(y_{t} \mid x, y_{<t}) \right],
\end{equation}
where $D$ represents the training set, $x$ and $y$ are the input and output sequences, respectively, $\theta$ denotes the trainable parameters of the LLM, $y_{t}$ is the $t$-th token in the output sequence, and $y_{<t}$ represents all preceding output tokens. \model is optimized to predict the CoT solution based on the provided database information and the corresponding question.

\section{Experiments}
In this section, we comprehensively evaluate the performance of \model by comparing it with leading LLMs.

\subsection{Experimental Setup}
\subsubsection{Evaluation Datasets} \textbf{Standard Benchmarks.} Spider~\cite{Yu2018@spider} and BIRD~\cite{Li2023@bird} are widely used in prior text-to-SQL studies to evaluate the cross-domain text-to-SQL capabilities of models. Cross-domain refers to the fact that the databases in the different data splits (train/dev/test) have no overlap. For Spider, we use its dev and test sets, containing 1,034 and 2,147 data samples, respectively. For BIRD, due to the hidden nature of its test set, we only use its dev set, which consists of 1,534 data samples.

\noindent\textbf{Challenging Domain-specific Benchmarks.} Spider2.0~\cite{Lei2024@spider2.0} is a new challenging benchmark focusing on the real-world enterprise text-to-SQL scenario. It contains highly complex SQL queries (often exceeding 100 lines), extremely long contexts (\emph{e.g.}, large database schemas and massive external knowledge), and multiple SQL dialects (\emph{e.g.}, BigQuery, Snowflake, SQLite, and DuckDB). Since \dataset is constructed using the SQLite dialect, we extract the SQLite portion of Spider2.0 for evaluation, containing 135 data samples. We name this subset as Spider2.0-SQLite\footnote{Our proposed data synthesis method can generate data for most mainstream SQL dialects (\emph{e.g.}, PostgreSQL, MySQL, SQL Server, BigQuery, etc.) because LLMs largely have seen these dialects in their pre-training corpora. We select SQLite as the default for this study because many classical text-to-SQL benchmarks use SQLite to host their databases, simplifying database deployment.}. Additionally, ScienceBenchmark~\cite{Zhang2023@sciencebenchmark}\footnote{The databases provided by ScienceBenchmark are originally hosted in PostgreSQL. We manually convert them to SQLite to facilitate evaluations.} and EHRSQL~\cite{Lee2022@ehrsql}\footnote{Many natural language questions in EHRSQL are unanswerable and several SQL queries return empty execution results. We remove these data samples from EHRSQL to ensure a consistent and reliable evaluation.} are two challenging benchmarks designed to evaluate text-to-SQL models in professional database domains. Specifically, ScienceBenchmark includes databases from the domains of research policymaking, astrophysics, and cancer research. EHRSQL, on the other hand, focuses on medical text-to-SQL applications. ScienceBenchmark and EHRSQL contain 299 and 1,008 evaluation samples, respectively. Since \model does not use any training data from these domain-specific benchmarks during fine-tuning, this evaluation can be treated as a zero-shot domain generalization scenario.

\noindent\textbf{Robustness Benchmarks.} Spider-DK~\cite{Gan2021@spider-dk}, Spider-Syn~\cite{Gan2021@spider-syn}, and Spider-Realistic~\cite{Deng2021@STRUG} are three widely-adopted robustness benchmarks. Spider-DK tests the model's ability to understand implicit domain knowledge in natural language questions. Spider-Syn and Spider-Realistic modify questions in Spider's development set to replace explicit mentions of column names with their synonyms. These designs can mimic real-world users' questions, enabling the evaluation of model robustness in practical text-to-SQL applications. Specifically, Spider-DK, Spider-Syn, and Spider-Realistic offer 535, 1,034, and 508 samples for evaluation.

\subsubsection{Evaluation Metrics}
Following prior work, we use execution accuracy (EX)~\cite{Yu2018@spider} and test-suite accuracy (TS)~\cite{Zhong2020@test-suites} as our evaluation metrics. EX measures whether the predicted SQL queries yield the same execution results as the gold SQL queries on a single database. TS extends EX's evaluation to multiple test-suite databases. Notably, only the Spider (dev), Spider-Syn, and Spider-Realistic datasets provide test-suite databases; therefore, these datasets are evaluated using TS, while the others are evaluated using EX. For both metrics, higher values indicate better performance.

\subsubsection{Implementation Details}\label{sec:impl_details}
\model-7B/14B/32B is built on the Qwen2.5-Coder-7B/14B/32B-Instruct models, a series of advanced code language models pre-trained and instruction-tuned on 92 programming languages. Specifically, \model-7B and \model-14B are fully fine-tuned, while \model-32B is fine-tuned using a parameter efficient technique, low-rank adaptation (LoRA)~\cite{Hu2022@lora}, due to limited GPU computational resources. For LoRA, we set $r = 256$ and $\alpha = 512$, integrating adapters into the \texttt{q\_proj}, \texttt{k\_proj}, and \texttt{v\_proj} layers of the model, while keeping the original weights unchanged. We use the AdamW optimizer~\cite{Loshchilov2019@adamw} with parameters $\beta_{1} = 0.9$, $\beta_{2} = 0.95$, and $\epsilon = 10^{-8}$ to optimize the training objective. The peak learning rates are set to $2e^{-5}$, $4e^{-6}$ and $2e^{-4}$ for \model-7B, \model-14B, and \model-32B, respectively. We use a learning rate schedule with a linear warmup for the initial 5\% of training, followed by cosine decay to 10\% of the peak rate. In addition, the batch size, number of epochs, context length, weight decay, and gradient clipping are uniformly set to 512, 2, 8192, 0.1, and 1.0, respectively. To optimize GPU memory usage during training, we leverage the DeepSpeed-Zero framework~\cite{Rajbhandari2020@deepspeed} with bfloat16 mixed precision. In practice, training \model-7B/14B/32B requires approximately 6, 12, and 20 days, respectively, on a single machine equipped with 8 NVIDIA A800-SXM4-80GB GPUs. We also provide LoRA versions of \model-7B and \model-14B (referred to as \model-7B-LoRA and \model-14B-LoRA), using the same LoRA hyperparameters as \model-32B. This enables a direct comparison between full and LoRA-based fine-tuning approaches.

\subsubsection{Environments} All experiments are conducted on two GPU servers, each equipped with 8 NVIDIA A800-SXM4-80GB GPUs, an Intel(R) Xeon(R) Platinum 8336C CPU, and 2TB of RAM. For LLM training, we utilize PyTorch~\cite{Ansel2024@pytorch2} 2.1.0 and DeepSpeed~\cite{Rajbhandari2020@deepspeed} 0.10.3, while vLLM~\cite{Kwon2023@vllm} 0.6.3 is employed for LLM inference.

\begin{table*}[t]
    \centering
    \caption{Main results on 9 datasets (\%). The best results are highlighted in bold. ``DSC'' is the abbreviation of ``DeepSeek-Coder''.}
    \label{tab:main_results}
    \setlength{\tabcolsep}{4pt}
    \footnotesize
    \renewcommand{\arraystretch}{0.9}
    \begin{tabular}{l|cccccc|cccccc|cccccc|cc}
       \toprule
       \multirow{2}{*}{\textbf{LLM}} & \multicolumn{2}{c}{\textbf{Spider}} & \multicolumn{2}{c}{\textbf{Spider}} & \multicolumn{2}{c|}{\textbf{BIRD}} & \multicolumn{2}{c}{\textbf{Spider2.0-}} & \multicolumn{2}{c}{\textbf{Science}} & \multicolumn{2}{c|}{\multirow{2}{*}{\textbf{EHRSQL}}} & \multicolumn{2}{c}{\textbf{Spider-}} & \multicolumn{2}{c}{\textbf{Spider-}} & \multicolumn{2}{c|}{\textbf{Spider-}} & \multicolumn{2}{c}{\multirow{2}{*}{\textbf{Average}}}\\
       & \multicolumn{2}{c}{\textbf{(dev)}} & \multicolumn{2}{c}{\textbf{(test)}} & \multicolumn{2}{c|}{\textbf{(dev)}} & \multicolumn{2}{c}{\textbf{SQLite}} & \multicolumn{2}{c}{\textbf{Benchmark}} & & & \multicolumn{2}{c}{\textbf{DK}}& \multicolumn{2}{c}{\textbf{Syn}} & \multicolumn{2}{c|}{\textbf{Realistic}} &  \\
       \midrule
        & Gre & Maj & Gre & Maj & Gre & Maj & Gre & Maj & Gre & Maj & Gre & Maj & Gre & Maj & Gre & Maj & Gre & Maj & Gre & Maj \\
       \midrule
       \multicolumn{21}{c}{\centering \textbf{Closed-source LLMs (as a reference)}} \\
       \midrule
       GPT-4o-mini & 70.4 & 71.0 & 82.4 & 83.7 & 58.8 & 61.5 & 5.9 & 7.4 & 51.8 & 52.5 & 37.9 & 43.1 & 73.3 & 74.4 & 60.5 & 61.6 & 64.4 & 66.7 & 56.2 & 58.0\\
       GPT-4-Turbo & 72.4 & 72.2 & 83.4 & 84.2 & 62.0 & 63.6 & 11.9 & 11.9 & 59.2 & 59.5 & 43.1 & 44.8 & 72.3 & 72.1 & 62.9 & 63.5 & 67.5 & 68.3 & 59.4 & 60.0\\
       GPT-4o & 70.9 & 70.7 & 83.2 & 84.9 & 61.9 & 64.0 & 14.8 & 15.6 & 55.5 & 56.2 & 44.9 & 45.5 & 72.9 & 73.5 & 59.6 & 62.3 & 66.5 & 66.7 & 58.9 & 59.9\\
       \midrule
       \multicolumn{21}{c}{\textbf{Open-source LLMs (\textasciitilde 7B)}} \\
       \midrule
       DSC-6.7B-Instruct & 63.2 & 63.2 & 70.5 & 73.2 & 43.1 & 48.0 & 3.0 & 3.7 & 40.8 & 45.5 & 28.6 & 33.9 & 60.9 & 64.1 & 49.9 & 51.7 & 58.7 & 58.9 & 46.5 & 49.1\\
       Qwen2.5-Coder-7B-Instruct & 73.4 & 77.1 & 82.2 & 85.6 & 50.9 & 61.3 & 1.5 & 2.2 & 45.2 & 51.2 & 24.3 & 36.9 & 67.5 & 73.6 & 63.1 & 66.9 & 66.7 & 70.5 & 52.8 & 58.4 \\
       Qwen2.5-7B-Instruct & 65.4 & 68.9 & 76.8 & 82.6 & 46.9 & 56.4 & 1.5 & 1.5 & 38.5 & 47.5 & 20.9 & 32.1 & 63.7 & 71.8 & 54.2 & 60.0 & 56.7 & 63.6 & 47.2 & 53.8 \\
       OpenCoder-8B-Instruct & 59.5 & 59.5 & 68.3 & 70.1 & 37.5 & 45.3 & 0.7 & 0.7 & 39.8 & 45.5 & 21.9 & 29.9 & 62.6 & 64.7 & 46.0 & 46.1 & 49.0 & 49.4 & 42.8 & 45.7 \\
       Meta-Llama-3.1-8B-Instruct & 61.8 & 67.7 & 72.2 & 78.5 & 42.0 & 53.1 & 3.0 & 2.2 & 36.8 & 43.1 & 24.6 & 33.7 & 62.6 & 69.9 & 53.1 & 59.3 & 57.5 & 61.0 & 46.0 & 52.1 \\
       Granite-8B-Code-Instruct & 58.5 & 59.2 & 64.9 & 68.6 & 27.6 & 32.5 & 0.7 & 0.7 & 29.4 & 31.4 & 16.0 & 22.6 & 50.7 & 54.4 & 45.0 & 46.8 & 48.8 & 49.4 & 38.0 & 40.6 \\
       Granite-3.1-8B-Instruct & 58.3 & 65.0 & 69.8 & 75.3 & 36.0 & 47.2 & 1.5 & 1.5 & 36.8 & 47.5 & 19.6 & 32.3 & 60.0 & 66.5 & 47.7 & 53.8 & 46.5 & 57.1 & 41.8 & 49.6 \\
       \midrule
       \textbf{\model-7B-LoRA} & 79.9 & 79.3 & 85.6 & 87.1 & 61.5 & \textbf{66.6} & 8.9 & \textbf{10.4} & \textbf{50.2} & 53.8 & \textbf{39.4} & \textbf{45.6} & 74.2 & 76.6 & 66.2 & 68.6 & 76.0 & 75.2 & 60.2 & 62.6 \\
       \textbf{\model-7B} & \textbf{81.2} & \textbf{81.6} & \textbf{87.9} & \textbf{88.9} & \textbf{63.9} & 66.1 & \textbf{10.4} & \textbf{10.4} & \textbf{50.2} & \textbf{55.9} & 34.9 & 40.0 & \textbf{76.1} & \textbf{77.8} & \textbf{69.7} & \textbf{69.6} & \textbf{76.2} & \textbf{78.0} & \textbf{61.2} & \textbf{63.1} \\
       \midrule
       \multicolumn{21}{c}{\textbf{Open-source LLMs (14B-32B)}} \\
       \midrule
       Qwen2.5-Coder-14B-Instruct & 78.1 & 80.6 & 86.6 & 88.0 & 61.5 & \textbf{66.1} & 5.9 & 4.4 & 52.2 & 54.2 & 31.6 & 35.5 & 73.6 & \textbf{77.8} & 68.2 & 69.3 & 76.2 & 74.2 & 59.3 & 61.1 \\
       Qwen2.5-14B-Instruct & 66.5 & 69.7 & 82.0 & 84.0 & 56.7 & 62.1 & 5.2 & 10.4 & 51.2 & 56.2 & 28.8 & 35.2 & 72.3 & 74.0 & 58.1 & 60.7 & 62.4 & 65.2 & 53.7 & 57.5\\
       Starcoder2-15B-Instruct & 65.8 & 67.6 & 73.0 & 74.0 & 38.5 & 42.6 & 0.7 & 3.0 & 25.8 & 29.8 & 16.8 & 22.6 & 66.5 & 68.2 & 49.4 & 52.4 & 56.7 & 61.0 & 43.7 & 46.8 \\
       DSC-V2-Lite-In. (16B, MoE) & 68.0 & 70.0 & 77.9 & 79.6 & 44.6 & 51.8 & 3.0 & 5.2 & 39.1 & 45.8 & 23.9 & 32.4 & 63.7 & 67.5 & 55.6 & 57.9 & 61.8 & 64.0 & 48.6 & 52.7 \\
       Granite-20B-Code-Instruct & 65.7 & 63.6 & 74.1 & 72.9 & 34.0 & 40.5 & 0.0 & 0.0 & 37.5 & 40.1 & 23.5 & 26.9 & 62.2 & 63.9 & 52.3 & 54.3 & 55.7 & 56.3 & 45.0 & 46.5 \\
       Codestral-22B  & 66.7 & 67.5 & 78.6 & 81.0 & 52.7 & 56.8 & 5.2 & 5.9 & 48.5 & 54.2 & 37.8 & 40.4 & 69.9 & 72.7 & 55.2 & 59.4 & 62.6 & 64.8 & 53.0 & 55.9 \\
       \midrule
       \textbf{\model-14B-LoRA} & 80.9 & 80.9 & 88.0 & 87.8 & 63.6 & 65.6 & \textbf{11.1} & 11.9 & \textbf{58.5} & 55.2 & 38.6 & \textbf{44.4} & \textbf{76.4} & 76.8 & \textbf{70.7} & \textbf{72.1} & \textbf{77.0} & \textbf{79.1} & \textbf{62.8} & 63.8 \\
       \textbf{\model-14B} & \textbf{81.4} & \textbf{82.0} & \textbf{88.3} & \textbf{88.3} & \textbf{64.2} & 65.9 & 10.4 & \textbf{13.3} & 56.9 & \textbf{56.9} & \textbf{39.9} & 43.6 & 72.9 & 74.8 & 69.0 & 72.0 & 76.4 & 78.5 & 62.2 & \textbf{63.9} \\
       \midrule
       \multicolumn{21}{c}{\textbf{Open-source LLMs ($\geq$ 32B)}} \\
       \midrule
       Qwen2.5-Coder-32B-Instruct & 77.7 & 77.9 & 87.5 & 88.0 & 64.5 & 67.0 & 5.9 & 11.9 & 54.8 & 56.5 & 36.4 & 43.3 & \textbf{78.3} & \textbf{78.1} & \textbf{69.9} & 70.5 & 72.4 & 74.8 & 60.8 & 63.1\\
       Qwen2.5-32B-Instruct & 71.9 & 73.6 & 84.9 & 86.1 & 62.0 & 64.7 & 7.4 & 8.9 & 50.5 & 54.5 & 33.6 & 41.4 & 73.1 & 76.1 & 64.0 & 66.0 & 66.5 & 68.1 & 57.1 & 59.9\\
       DSC-33B-Instruct & 66.0 & 68.5 & 74.3 & 76.5 & 49.2 & 55.9 & 7.4 & 4.4 & 44.5 & 52.2 & 31.4 & 35.4 & 69.0 & 71.4 & 53.5 & 57.4 & 59.1 & 63.2 & 50.5 & 53.9\\
       Granite-34B-Code-Instruct & 69.9 & 70.0 & 74.4 & 77.0 & 33.8 & 41.3 & 0.7 & 1.5 & 40.1 & 40.1 & 23.8 & 29.9 & 64.7 & 70.7 & 55.6 & 59.8 & 60.0 & 59.6 & 47.0 & 50.0 \\
       Mixtral-8x7B-In. (47B, MoE) & 54.4 & 59.0 & 67.8 & 74.1 & 35.3 & 42.9 & 2.2 & 2.2 & 29.4 & 34.8 & 21.5 & 31.4 & 55.3 & 60.4 & 42.1 & 48.8 & 48.0 & 53.3 & 39.6 & 45.2 \\
       Meta-Llama-3.1-70B-Instruct & 72.3 & 71.0 & 84.3 & 85.9 & \textbf{65.1} & \textbf{67.4} & 3.0 & 3.7 & 55.2 & 56.2 & 37.4 & 41.4 & 75.1 & \textbf{78.1} & 61.7 & 63.1 & 64.0 & 65.6 & 57.6 & 59.2\\
       Qwen2.5-72B-Instruct & 73.9 & 72.1 & 84.0 & 85.7 & 60.3 & 63.6 & 9.6 & 14.8 & 52.8 & 58.2 & 35.0 & 41.2 & 76.4 & 77.6 & 64.1 & 64.3 & 70.1 & 68.5 & 58.5 & 60.7\\
       DeepSeek-V3 (671B, MoE) & 73.1 & 73.5 & 85.5 & 85.8 & 63.2 & 63.8 & \textbf{12.6} & \textbf{15.6} & 56.2 & 57.9 & \textbf{43.2} & 43.5 & 72.9 & 73.8 & 64.4 & 65.1 & 67.9 & 66.9 & 59.9 & 60.7\\
       \midrule
       \textbf{\model-32B} & \textbf{80.9} & \textbf{80.9} & \textbf{87.6} & \textbf{89.8} & 64.5 & 67.0 & 11.9 & 13.3 & \textbf{57.2} & \textbf{58.5} & 42.4 & \textbf{46.8} & 76.1 & 77.6 & 69.7 & \textbf{72.1} & \textbf{78.1} & \textbf{77.2} & \textbf{63.2} & \textbf{64.8} \\
    \bottomrule
    \end{tabular}
\end{table*}

\subsubsection{Baselines}
We compare \model with a wide range of LLMs, including closed-source LLMs such as GPT-4o-mini-2024-07-18~\cite{openai2024@gpt4o-mini}, GPT-4o-2024-11-20~\cite{openai2024@gpt4o}, and GPT-4-Turbo-2024-04-09~\cite{openai2024@gpt4-turbo}, as well as open-source LLMs like DeepSeek-V3~\cite{deepseek2024@deepseek-v3}, DeepSeek Coder~\cite{Guo2024@deepseek-coder, deepseek@deepseek-coder-v2}, Qwen2.5~\cite{Yang2024@qwen2.5}, Qwen2.5 Coder~\cite{hui2024@qwen2.5-coder}, LLama-3.1~\cite{Dubey2024@llama3}, Granite~\cite{Mishra2024@granite-coder}, Mixtral~\cite{Jiang2024@mixtral}, OpenCoder~\cite{Huang2024@opencoder}, and Starcoder2~\cite{Lozhkov2024@starcoder2}. These models span a wide range, from closed-source to open-source, small-scale (6.7B) to large-scale (671B), general-purpose to code-specific, and dense architectures to Mixture-of-Expert (MoE) designs. For closed-source LLMs with undisclosed parameter sizes, their performance serves as a point of reference rather than a direct comparison under controlled parameter size conditions. Additionally, due to the prohibitively high inference costs, reasoning models such as OpenAI o1~\cite{openai2024@o1} and DeepSeek-R1~\cite{deepseek2025@r1} are not considered.

\subsubsection{Inference Strategy}
During LLM inference, we explore greedy decoding and sampling. Greedy decoding, with a temperature of 0, ensures deterministic responses. Sampling, at a temperature of 0.8, introduces creativity and diversity, generating 8 candidate responses per sample. Then, we extract SQL queries from these candidates and perform majority voting based on execution results. The final response is selected from the group with the most votes.

\subsection{Main Results}\label{sec:main_results}
The evaluation results are shown in Table~\ref{tab:main_results}, where ``Gre'' and ``Maj'' denote greedy decoding and majority voting results, respectively. At each model scale, we compare \model with open-source LLMs of similar or larger size. Our findings are listed as follows:

\textbf{Synthetic data significantly enhances the base model's text-to-SQL capabilities.} A comparison between the Qwen2.5-Coder models and \model demonstrates that fine-tuning with \dataset leads to improved performance across most datasets. Specifically, under the greedy decoding strategy, \model-7B achieves an average improvement of +8.4\% (from 52.8\% to 61.2\%) over its base model, Qwen2.5-Coder-7B-Instruct. Similarly, \model-14B and \model-32B show average improvements of 2.9\% (from 59.3\% to 62.2\%) and 2.4\% (from 60.8\% to 63.2\%) over their base models, Qwen2.5-Coder-14B-Instruct and Qwen2.5-Coder-32B-Instruct, respectively. Notably, \model-7B sets a new standard for 7B-scale LLMs in this domain. Furthermore, \model-14B and \model-32B represent the new state-of-the-art text-to-SQL capabilities.

\textbf{\model consistently demonstrates leading performance on standard benchmarks, including Spider (dev), Spider (test), and BIRD (dev).} Notably, \model-7B model attains an accuracy of 87.9\% (greedy decoding) on Spider's test set, surpassing the best publicly available method on Spider's leaderboard\footnote{Spider's leaderboard can be found at \url{https://yale-lily.github.io/spider}. The top-ranked method, MiniSeek, is excluded from comparisons as it is not publicly available.}, DAIL-SQL + GPT-4 + Self-Consistency (86.6\%)~\cite{Gao2024@dailsql}, by 1.3\%. Additionally, by employing a simple majority voting strategy, \model's performance on the Spider test set improves to 88.9\%, 88.3\%, and 89.8\% for the 7B, 14B, and 32B model scales, respectively. In addition, we observe that increasing the model scale does not yield consistent performance gains on Spider. This is likely due to Spider being a relatively simple benchmark, where smaller-scale but powerful models already achieve near-optimal performance. In contrast, as BIRD is more challenging than Spider, we can observe slight performance improvements as the scale of \model increases. Notable, \model-32B achieves 67.0\% (major voting) accuracy on BIRD's development set, making it competitive with Distillery + GPT-4o (67.2\%)~\cite{Maamari2024@Distillery}, which fine-tunes GPT-4o on BIRD's training set.

\textbf{\model demonstrates exceptional domain generalization capabilities on domain-specific benchmarks, including Spider2.0-SQLite, EHRSQL, and ScienceBenchmark.} On the most challenging benchmark, Spider2.0-SQLite, despite having limited model parameters, \model still shows comparable accuracy to much larger models. Notably, for 7B-scale baseline LLMs, their accuracy is only in the range from 0.7\% to 3.7\%, while \model-7B achieves 10.4\% (greedy decoding) accuracy, underscoring the potential of smaller models to handle complex text-to-SQL problems. Then, on EHRSQL and ScienceBenchmark, \model consistently outperforms similar scale competitors and matches the performance of leading LLMs. Remarkably, on EHRSQL, which involves medical domain databases, \model-32B achieves the best performance, 46.8\% (major voting), outperforming GPT-4o (45.5\% in major voting) by 1.3\%. These results underscore \model's capability to generalize effectively across professional domain databases without necessitating extensive domain-specific fine-tuning.

Interestingly, some open-source LLMs (\emph{e.g.}, Qwen2.5-Coder-32B-Instruct and Meta-Llama-3.1-70B-Instruct) significantly outperform closed-source LLMs on standard datasets. However, their performance advantage diminishes on domain-specific datasets. This is likely because these open-source models incorporated the training sets of popular text-to-SQL benchmarks (\emph{e.g.}, Spider and BIRD) during pre-training or fine-tuning. While this enables strong performance on familiar datasets, their effectiveness decreases when encountering out-of-domain datasets. In contrast, \model demonstrates consistently strong performance, excelling on both standard and domain-specific benchmarks.

\begin{table*}[t]
    \centering
    \caption{Results of ablation studies (\%). All scores are reported under greedy decoding. FT means ``fine-tuning''. }
    \label{tab:ablation}
    \setlength{\tabcolsep}{4pt}
    \footnotesize
    \renewcommand{\arraystretch}{0.9}
    \begin{tabular}{l|ccc|ccc|ccc}
       \toprule
        & \textbf{Spider} & \textbf{Spider} & \textbf{BIRD} & \textbf{Spider2.0-} & \textbf{Science} & \multirow{2}{*}{\textbf{EHRSQL}}  & \textbf{Spider-} & \textbf{Spider-} & \textbf{Spider-} \\
       & \textbf{(dev)} & \textbf{(test)} & \textbf{(dev)} & \textbf{SQLite} & \textbf{Benchmark} & & \textbf{DK}& \textbf{Syn} & \textbf{Realistic} \\
       \midrule
       Qwen2.5-Coder-7B-Instruct (base model) & 73.4 & 82.2 & 50.9 & 1.5 & 45.2 & 24.3 & 67.5 & 63.1 & 66.7 \\
       \midrule
       FT w/ \dataset & 74.6 & 83.5 & 59.9 & 9.6 & 48.5 & \textbf{37.2} & 71.6 & 61.6 & 70.3 \\
       FT w/ CoT-enhanced Spider + BIRD & 80.3 & 86.6 & 59.6 & 5.9 & 49.5 & 26.0 & 72.0 & \textbf{70.0} & \textbf{76.4} \\
       FT w/ original Spider + BIRD & 76.9 & 80.7 & 55.1 & 3.0 & 43.8 & 31.4 & 65.8 & 65.9 & 71.9 \\
       FT w/ \dataset + CoT-enhanced Spider + BIRD (\textbf{\model-7B}) & \textbf{81.2} & \textbf{87.9} & \textbf{63.9} & \textbf{10.4} & \textbf{50.2} & 34.9 & \textbf{76.1} & 69.7 & 76.2\\
    \bottomrule
    \end{tabular}
\end{table*}

\textbf{\model demonstrates strong robustness on Spider-DK, Spider-Syn, and Spider-Realistic.} On Spider-Syn and Spider-Realistic, \model outperforms baselines in most cases, demonstrating robustness to synonym substitution—crucial for real-world scenarios where users may use similar terms for tables or columns. However, on Spider-DK, we note that \model-14B and \model-32B underperform compared to their base models (\emph{i.e.}, Qwen2.5 Coder) on Spider-DK. We attribute this to the fact that Spider-DK requires the text-to-SQL model to have a deep understanding of implicit knowledge (\emph{e.g.}, commonsense knowledge) hidden in the questions. The base models, trained on extensive corpora, likely possess a stronger foundation in such knowledge, enabling them to perform better on Spider-DK. These findings provide valuable insights for further refining our data synthesis framework. For example, we could introduce a new language style to guide the LLM in synthesizing questions that incorporate commonsense knowledge.

\textbf{Smaller models benefit more from full fine-tuning, while LoRA is sufficient for larger models.} For the 7B model, full fine-tuning (\model-7B) outperforms the LoRA-based version (\model-7B-LoRA) in most cases. However, this advantage becomes less pronounced with larger models (see \model-14B vs. \model-14B-LoRA). We hypothesize that larger models possess stronger inherent capabilities, so updating only a subset of parameters with LoRA is sufficient for effective adaptation and may also help mitigate overfitting compared to full fine-tuning.

Finally, it is important to note that \model's performance can be further enhanced by incorporating additional text-to-SQL techniques, such as question rephrasing~\cite{AlphaSQL2025@li}, schema linking~\cite{Li2023@resdsql, Pourreza2023@dinsql}, SQL revision~\cite{Talaei2024@chess, Gao2024@xiyansql}, and SQL selection~\cite{Pourreza2024@chasesql, Lee2025@mcssql}. Recent open-source text-to-SQL frameworks like CHASE-SQL~\cite{Pourreza2024@chasesql} and Alpha-SQL~\cite{AlphaSQL2025@li} leverage multi-stage pipelines and integrate a range of such techniques. In contrast, our work isolates and evaluates the core text-to-SQL capability of a single LLM in a single-step inference setup. To ensure fair evaluation, we focus on this setting and do not compare \model directly with multi-stage frameworks.

\begin{figure}[t]
    \centering
    \includegraphics[width=0.96\linewidth]{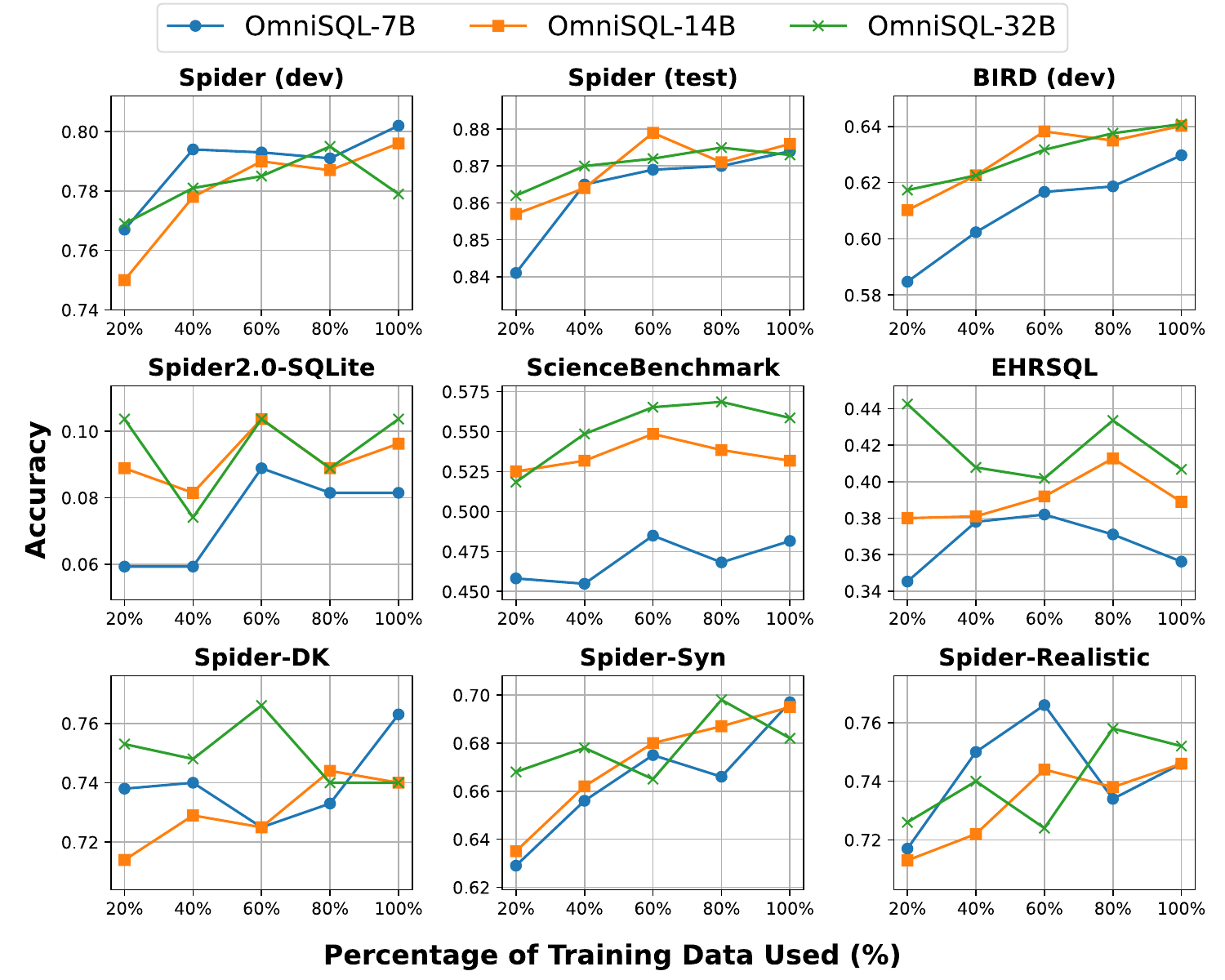}
	\caption{\label{fig:data_scale} Performance changes (measured by accuracy; higher is better) with increasing training data.}
\end{figure}

\subsection{Ablation studies}\label{sec:ablation}
\subsubsection{Ablations on Synthetic Data}\label{sec:ablation_syn_data}
To assess the impact of our synthetic data, we conduct two ablation studies. First, we fine-tune the base model using only \dataset. As shown in Table~\ref{tab:ablation} (FT w/ \dataset), this approach yields substantial improvements across eight datasets compared to the base model, including notable gains of 9.0\% on the BIRD dev set and 12.9\% on EHRSQL, underscoring the effectiveness of \dataset. Second, to further demonstrate the contribution of \dataset, we exclude it from the training set of \model and fine-tune the base model solely on CoT-enhanced Spider and BIRD. By comparing ``FT w/ CoT-enhanced Spider + BIRD'' with ``FT w/ \dataset + CoT-enhanced Spider + BIRD'' in Table~\ref{tab:ablation}, we observe consistent performance drops across seven datasets when \dataset is omitted, highlighting its key role in complementing gaps left by human-annotated data.

\subsubsection{Ablations on CoT Synthesis}
We conduct ablation studies to assess the impact of synthesizing CoT solutions (the fourth step in our pipeline). Specifically, we fine-tune Qwen2.5-Coder-7B-Instruct on the original training sets of Spider and BIRD, which contain only gold SQL queries as labels. As shown in Table~\ref{tab:ablation}, removing CoT solutions from the training sets leads to notable performance drops on nearly all datasets (except EHRSQL), as seen by comparing the rows ``FT w/ CoT-enhanced Spider + BIRD'' and ``FT w/ original Spider + BIRD''. The improvements from CoT can be attributed to both enhanced reasoning in the LLM and the correction of mislabeled samples during CoT synthesis, thus improving data quality.

\subsubsection{Ablations on Training Data Scale}
We assess the effect of training data scale by fine-tuning models on randomly sampled subsets (20\%, 40\%, 60\%, 80\%, and 100\%) for one epoch and reporting greedy decoding results in Figure~\ref{fig:data_scale}. The results show that model performance consistently improves with larger training data. Notably, the upward trend persists even with the full dataset, suggesting further gains are possible with more data.

\subsubsection{Ablations on CoT Synthesis Stage}\label{sec:ablation_cot}
As discussed in Section~\ref{sec:cot_synthesis}, we empirically observe that adding the CoT synthesis step helps correct prior errors and improves SQL quality. To validate this, we sampled 5,000 cases from \dataset where SQL execution results differed before and after CoT synthesis. GPT-4o is then used to judge which SQL query better matches the question. In 4,699 out of 5,000 cases (93.98\%), the SQL after CoT synthesis is preferred, confirming that this step substantially enhances the quality and reliability of the synthetic data.

\begin{table}[t]
    \centering
    \footnotesize
    \caption{Results of data augmentation methods (\%). $\dagger$ denotes methods using the EX metric on Spider (dev). Columns show results without synthetic data (w/o), with synthetic data (w/), and the improvement ($\Delta$) from synthetic data.}
    \label{tab:data_aug_results}
    \setlength{\tabcolsep}{4pt}
    \renewcommand{\arraystretch}{0.9}
    \begin{tabular}{l|ccc|ccc}
       \toprule
       \textbf{Method} & \multicolumn{3}{c}{Spider (dev)} & \multicolumn{3}{c}{BIRD (dev)} \\
       & w/o & w/ & $\Delta$ & w/o & w/ & $\Delta$ \\
       \midrule
       DT-Fixup+Syn data$^{\dagger}$~\cite{Yang2021@hierarchical-DA} & 74.6 & 76.1 & +1.5 & - & - & - \\
       T5-3B+PICARD+Syn data$^{\dagger}$~\cite{Hu2023@importance-DA} & 79.3 & 81.4 & +2.1 & - & - & - \\
       Sense-13B~\cite{Yang2024@sense} & 79.4 & 82.9 & +3.5 & 51.6 & 52.9 & +1.3 \\
       \midrule
       \model-7B (greedy decoding) & 76.9 & 81.2 & \textbf{+4.3} & 55.1 & 63.9 & \textbf{+8.8} \\
    \bottomrule
  
    \end{tabular}
\end{table}

\subsection{Comparison with Data Augmentation}
Finally, we compare our data synthesis method with 3 recent and representative data augmentation methods. To ensure a fair comparison, we focus on the performance improvements brought by synthetic data rather than absolute accuracy, as different methods use varying base models. As shown in Table~\ref{tab:data_aug_results}, our method achieves significant accuracy gains on both the Spider and BIRD development sets, outperforming existing data augmentation methods by a large margin. Notably, while Sense~\cite{Yang2024@sense} also uses LLMs for data synthesis, it achieves only a modest improvement (+1.3\%) on BIRD. In contrast, our approach achieves a substantial +8.8\% improvement on BIRD, highlighting its effectiveness.

\section{Conclusion}
This paper introduces a novel framework for text-to-SQL data synthesis, which breaks the process into four sequential, controllable steps. Using this framework, we introduce \dataset, a new text-to-SQL dataset containing 2.5 million high-quality samples. Comprehensive statistical analysis and quality evaluation demonstrate the superior quality, diversity, and complexity of \dataset. Based on this dataset, we introduce a new text-to-SQL model, \model. Extensive evaluations across nine benchmarks show that \model substantially outperforms baseline LLMs of similar scale and matches the performance of state-of-the-art models, despite using far fewer parameters. We believe that releasing both \dataset and \model will provide valuable resources and drive further advances in text-to-SQL research.

\begin{acks}
This work is supported by the National Key Research \& Development Plan of China (2023YFF0725100) and the National Natural Science Foundation of China (62322214, U23A20299, U24B20144, 62172424, 62276270). We also acknowledge the support of the Public Computing Cloud, Renmin University of China.
\end{acks}

\balance
\bibliographystyle{ACM-Reference-Format}
\bibliography{sample}

\end{document}